\documentclass[manuscript]{acmart}
\AtBeginDocument{%
  \providecommand\BibTeX{{%
    \normalfont B\kern-0.5em{\scshape i\kern-0.25em b}\kern-0.8em\TeX}}}

\setcopyright{acmcopyright}
\copyrightyear{2022}
\acmYear{2022}
\acmDOI{XXXXXXX.XXXXXXX}

\usepackage{subcaption} 

\usepackage{color}
\usepackage{xcolor}

\newcommand{\E}{\mathcal{E}}

\newtheorem{thm}{Theorem}[section]

\newtheorem{asum}[thm]{Assumption}

\usepackage{algorithm}
\usepackage{algorithmic}

\begin{document}

\title{Tabular and Deep Learning for the Whittle Index}

\author{Francisco Robledo Relaño$^*$}
\thanks{$^*$ Corresponding author. The authors are listed in reversed alphabetical order.}
\email{frrobledo96@gmail.com}
\orcid{0000-0003-1040-1513}
\affiliation{%
  \institution{UPV/EHU, Univ. of the Basque Country}
  \city{Donostia}
  \state{Gipuzkoa}
  \country{Spain}
  \postcode{20018}
}

\author{Vivek Borkar}
\email{borkar.vs@gmail.com}
\orcid{0000-0003-0756-5402}
\affiliation{%
    \institution{Indian Institute of Technology Bombay}
    \country{India}
}

\author{Urtzi Ayesta}
\email{urtzi.ayesta@irit.fr}
\orcid{0000-0003-1761-2313}
\affiliation{
    \institution{CNRS-IRIT, UPV/EHU and IKERBASQUE}
    \country{France and Spain}
}  

\author{Konstantin Avrachenkov}
\email{k.avrachenkov@inria.fr}
\orcid{0000-0002-8124-8272}
\affiliation{%
    \institution{INRIA Sophia Antipolis}
    \country{France}
}

\renewcommand{\shortauthors}{Francisco Robledo, et al.}

\begin{abstract}
The Whittle index policy is a heuristic that has shown remarkably good performance (with guaranteed asymptotic optimality) when applied to the class of problems known as Restless Multi-Armed Bandit Problems (RMABPs). In this paper we present QWI and QWINN,  { two reinforcement learning algorithms, respectively tabular and deep,  to learn the Whittle index for the total discounted criterion.}  The key feature is the use of two time-scales, a faster one to update  the state-action $Q$-values, and a relatively slower one to update the Whittle indices. In our main theoretical result we show that QWI, which is a tabular implementation,  converges to the real Whittle indices.
We then present QWINN, an adaptation of QWI algorithm using neural networks to compute the $Q$-values on the faster time-scale, which is able to extrapolate information from one state to another and scales naturally to large state-space environments. For QWINN, we show  that all local minima of the Bellman error are locally stable equilibria, which is the first result of its kind for DQN-based schemes.
Numerical computations show that QWI and QWINN converge  faster than the standard $Q$-learning algorithm, neural-network based approximate Q-learning and other state of the art algorithms. 
\end{abstract}



\maketitle

\section{Introduction}\label{sec_introduction}%
Markov decision processes (MDPs) provide a mathematical framework to model sequential decision problems. Formally, an MDP is a stochastic control process where the objective is to maximise a long-run payoff. At each time step, depending on the state and action taken by the MDP, the decision maker receives a reward and reaches a new random state. Due to their broad applicability, MDPs are found in many areas, including artificial intelligence, economics, communications and operations research.

Solving an MDP typically involves dynamic programming, a method often computationally prohibitive for realistically sized models. Consequently, significant attention has been devoted to classes of MDPs that are either analytically tractable or amenable to precise approximations. This paper focuses on one such class - Restless Multi-Armed Bandit Problems (RMABPs), initially introduced in \cite{whittle1988restless}. 
In an RMABP there are $N$  concurrent projects or bandit's arms. The decision maker knows the states of all arms and the reward in every state, and aims at maximizing the long-term reward. At every decision epoch, the decision maker activates $M<N$ arms, and the state of active and passive arms evolve stochastically. We note that RMABPs are a generalization of multi-arm bandit problems (MABP), in which only one arm can be activated at a time, and the state of passive arms does not evolve. It was shown by Gittins that the optimal solution to a MABP is an index policy, nowadays known as Gittins' index policy, see \cite{GGW11}. 

Over the years, RMABPs have gained significant traction, finding applications in diverse domains such as inventory routing, machine maintenance, healthcare systems, and networking. Solving RMABPs precisely is typically infeasible, except for simplified examples, and they are classified as PSPACE-hard problems \cite{PapTsi1999}. 
In \cite{whittle1988restless} Whittle developed a heuristic by  solving a relaxed version of the RMABP in which $M<N$ arms are activated on average. The resulting policy, known as Whittle index policy, relies on calculating Whittle index for each of the arms, and activating in every decision epoch the $M$ arms with the highest Whittle indices. It has been reported on numerous instances that Whittle index policy provides strikingly good performance, and it has been shown to be asymptotically optimal as the number of arms grows large \cite{weber1990index}.  As expected, Whittle index reduces to Gittins index when applied to an MABP.

In recent years there has been a surge in the interest in developing algorithms to learn the Whittle index in model-free setting. The first paper aimed at learning index policies is to the best of our knowledge,  \cite{duff1995q}, which considered an MABP and developed an algorithm that learns Gittins' indices. Regarding Whittle index, one of the first papers was \cite{fu2019towards}, which proposed an algorithm that did not converge to the correct values. {More recently, the authors of \cite{Gibson2021} focused on the resolution of the restless multi-armed bandit by iteratively obtaining a policy similar to the Whittle index policy, but not necessarily a precise value of the Whittle index themselves. Furthermore, the algorithm in \cite{Gibson2021} is a tabular only algorithm, making it unsuitable for large state spaces where the exploration of all states becomes impossible.}

Further contributions to the field include the work by \cite{Liu2010}, which established a foundation for the indexability concept in restless bandit problems and the associated optimality of Whittle indices for dynamic channel access. \cite{Hsu2018} showcased the adaptability of the Whittle index for stochastic scheduling within a restless multi-armed bandit framework. \cite{Verloop2016} offered insights into asymptotically optimal priority policies in settings involving both indexable and non-indexable bandits, illustrating the robustness of Whittle index-based heuristics. Moreover, \cite{NinoMora2020} provided an examination of threshold-indexability and highlighted the practical utility of the Whittle index for single-project control and as a heuristic policy.

In their exploration of the time-average criterion, Avrachenkov and Borkar \cite{Avrachenkov2022} devised a tabular algorithm that successfully converges to the Whittle index. The tabular setting with the time-average criterion was then extended to a Neural Network (NN) setting in \cite{Pagare2023}. A finite-time analysis of the scheme of \cite{Avrachenkov2022} with a neural network approximate has been also recently performed in \cite{xiong2023finite}. Killian et al. \cite{Killian2021} explore the online learning landscape for multi-action RMABs, presenting a Q-learning Lagrange policy algorithm tailored for restless multi-armed bandits with multiple discrete actions. Another notable mention is the NeurWIN algorithm \cite{Nakhleh2021}, a NN gradient-based reinforcement learning approach for solving RMABPs. NeurWIN, focused on maximizing discounted long-term rewards, does not, however, guarantee convergence to the true Whittle indices.

This paper introduces QWI and QWINN, two novel reinforcement learning algorithms, engineered to master the Whittle index in the realm of discounted rewards. Both algorithms leverage a dual time-scale strategy: a rapid cycle for updating state-action $Q$-values and a slower one for refining Whittle indices. Our principal theoretical finding asserts that QWI, with its tabular form, consistently aligns with the Whittle indices across various RMABPs. Additionally, we unveil QWINN, an evolution of QWI that integrates neural networks, drawing inspiration from DQN \cite{mnih2015human}. QWINN excels in large state spaces due to its neural network-driven extrapolation capabilities. We establish, for the first time to our knowledge, that QWINN ensures that local minima of the Bellman error function form stable equilibria in the differential equation limit of the DQN model. Thus, when QWINN starts within the attraction domain of such equilibria, convergence is highly probable. This interesting finding for DQN-based methods marks a significant advance in the field. Our numerical comparisons, pitting QWI and QWINN against established algorithms like standard $Q$-learning, DQN, and NeurWIN, demonstrate superior performance in both convergence rate and discounted reward optimization, with QWINN exhibiting remarkable proficiency in deriving accurate Whittle indices from limited data samples.

We compare numerically the performance of QWI and QWINN with several other algorithms, including the standard $Q$-learning algorithm, a vanilla implementation of DQN  and NeurWIN. 
{Unlike \cite{Wang2020} and \cite{ortner2012regret}, we dot not analyse the regret of our algorithms. Instead, we will study the convergence to the theoretical values of the Whittle indices and assess the performance of the algorithms over time as a function of the policy they have learned so far.}
Our results show that both QWI and QWINN outperform the other algorithms both in terms of the rate of convergence as well as the discounted reward. 
QWINN stands out in its ability to obtain good Whittle indices with far fewer samples than QWI, and  thanks to the extrapolation of the information performed by the neural networks, it is able to obtain good predictions of the indices of these states even with few or no samples. On the other hand, QWI, given enough samples in each state/action pair, is able to reliably converge to the correct values of the Whittle indices. 

\section{Restless Markovian Bandits}\label{sec_restless}
\subsection{Problem formulation and relaxation}\label{sec_prob_formulation}
In our study of Restless Markovian Bandits, we examine a Restless Multi-Armed Bandit Problem (RMABP) comprising $N$ arms. Each arm $i$, where $i=1, \ldots, N$, possesses a state space $\mathcal{S}^i$ of cardinality $|\mathcal{S}^i|$. The combined state space for all arms is denoted as $\mathcal{S} = \mathcal{S}^1 \times \cdots \times \mathcal{S}^N$ with cardinality $|\mathcal{S}|$. For each arm at time step $n$, $s_n^i$ represents the current state and $a_n^i$ the chosen action. The reward garnered by arm $i$ at this step is given by $r^i(s_n^i, a_n^i)$. We adopt the total discounted reward criterion with a discount factor $0<\gamma<1$. Active arms transition between states with probability $p(s^i_{n+1}| s^i_{n}, 1)$, while passive arms use a different transition probability $p(s^i_{n+1}| s^i_{n}, 0)$.

At each time step $n$, the control policy $\pi$ observes the state of all the arms, and activates $M$ arms ($a_n^i=1$), whereas the rest remain passive with $a_n^i=0$. The objective is to determine the optimal control policy $\pi^*$ that solves:
\begin{equation}\label{eq_discounted reward}
V_{\pi^*}(s_1,\dots,s_N) =\max_\pi    E_s \left[\sum_{n=1}^\infty \sum_{i=1}^N \gamma^n r^i \left(s_n^i, a_n^i\right) \right], 
\end{equation}
subject to:
\begin{equation}\label{eq_constraint}
    \sum_{i=1}^N a_n^i = M, \phantom{-} n \geq 0,
\end{equation}
where $V_{\pi^*}$ is known as the value function.  

Whittle's approach \cite{whittle1988restless} relaxes the constraint to be satisfied on average, leading to an unconstrained problem formulation: $E_s\left[\sum_{n=0}^\infty\sum_{i=1}^N \gamma^n a_n^i\right] \leq M/(1-\gamma)$. The problem then has an equivalent \emph{unconstrained} formulation:
\begin{equation}\label{eq_relaxed reward}
 \max_{\pi}   E_s \left[ 
    \sum_{n=1}^\infty \sum_{i=1}^N \gamma^n \left(
        r \left(s_n^i, a_n^i\right) + \lambda \left(1 - a_n^i\right)
    \right)
    \right]
\end{equation}
where $\lambda$ is a Lagrange multiplier associated with the constraint. We note that as $\lambda$ increases, we expect the passive action to become attractive in more states, and as a consequence, the multiplier $\lambda$ can be seen as a subsidy for passivity.

The key observation by Whittle is that problem~(\ref{eq_relaxed reward}) can be decomposed, and its  solution is obtained by combining the solution of $N$ independent problems. In other words, for each arm $i$ we need to solve the associated Bellman equation given by:

\begin{equation}\label{eq_value function}
    \begin{aligned}
        V_\pi^{i}(s)=  &\max_{a \in \{0,1\}} \left[Q^i(s,a)\right]
    \end{aligned}
\end{equation}
where
\begin{equation}\label{eq_q value teor}
    \begin{aligned}
        Q_\pi^i(s,a) =  a
    \left(
    r^i(s,1) + \gamma \sum_j p^i(j|s,1) V_\pi^i(j) 
    \right) + 
     (1 - a)
    \left(
    r^i(s,0) + \lambda + \gamma \sum_j p^i(j|s,0) V_\pi^i(j)
    \right)
    \end{aligned}
\end{equation}
The functions $V^i(s)$ and $Q^i(s,a)$ are known as the value function and the state-action function resp.\ for arm $i$. 

\subsection{Whittle index}\label{sec_whittleindex}

The optimal action in (\ref{eq_value function}) will be to activate a given arm $i$ if $r^i(k,1) + \gamma \sum_j p^i(j|k,1) V^i(j) > r^i (k,0) + \lambda + \gamma \sum_j p^i(j|k,0)V^i(j)$, while otherwise the optimal action will be to keep it passive. 
For a given $\lambda$, the optimal policy $\pi^*$ is then characterized by $\mathcal{S}(\lambda)$,  the set of states in which the optimal action is to activate. 
 
 The problem is deemed \emph{indexable} if the subset of states where activation is optimal decreases monotonically with increasing $\lambda$. For a specific state $k$, the Whittle index, $\lambda(k)$, equates the expected rewards for active and passive actions:
 \begin{equation}
r^i(k,1) + \gamma \sum_j p^i(j|k,1) V_{\pi^*}^i(j) = r^i (k,0) + \lambda(k) + \gamma \sum_j p^i(j|k,0)V_{\pi^*}^i(j).
\label{eq:cond-whittle}
 \end{equation}
 
It thus follows that the Whittle index characterizes the optimal solution to the relaxed problem (\ref{eq_relaxed reward}), which will simply activate all arms whose Whittle index is larger than $\lambda$.

 
 In his seminal paper, Whittle introduced the heuristic policy for the problem (\ref{eq_discounted reward}), known nowadays as Whittle index policy, which is  defined as the policy that at every time step $n$ activates the $M$ arms with the $M$ highest Whittle indices. As explained in the introduction, this heuristic has shown to have a close to optimal performance, and to be asymptotically optimal as the values of both $N$ and $M$ tend to infinity, see~\cite{weber1990index}~and~\cite{Verloop2016}.

\section{Learning the Whittle indices}\label{sec_QWI}
In this section we present QWI and QWINN,  the tabular and neural network-based algorithms to learn Whittle indices. Throughout this section we drop the dependence on the arm from the notation. 

\subsection{Tabular QWI algorithm}\label{sec_tabularQWI}
In this subsection, we delve into the Tabular QWI algorithm, a method designed to learn Whittle indices in a structured, iterative manner. The core idea is to iteratively adjust the Whittle index, refining our estimate of the state-action function until convergence. The Whittle index $\lambda(x)$ for a state $x$ can be implicitly derived from the equilibrium condition in Equation (\ref{eq:cond-whittle}):
\begin{equation}
\lambda(x): Q(x,1)-Q(x,0) = 0.
\label{eq:fixedpointQ}
\end{equation}
This is an implicit equation, as the state-action function $Q(\cdot,\cdot)$ depends on the value of the multiplier $\lambda(x)$. The Whittle index is thus defined by coupled equations (\ref{eq_q value teor}) and (\ref{eq:fixedpointQ}).

To find this equilibrium, we propose a method that alternates between estimating the state-action functions and updating the Whittle index. This process is conducted online, ensuring convergence to the actual Whittle index. The QWI algorithm employs stochastic approximation with multiple time scales (Chapter 6 in \cite{borkar2008stochastic}) to achieve this.

For each arm $i$, from the samples $(s_{n}, a_{n}, r_{n}, s_{n+1})$:
\begin{equation}\label{eq_q-learning with x}
    \begin{aligned}
        Q_{n+1}^{x}(s_n,a_n) =  (1-\alpha(n))Q_{n}^{x}(s_n,a_n) + \alpha(n)
         \left( (1-a_n)(r_0(s_n) + \lambda_n(x)) +  a_n r_1(s_n) +\gamma \max_{v \in \{0,1\}} Q_n^{x}(s_{n+1}, v) \right)
    \end{aligned}
\end{equation}
and
\begin{equation}
    \lambda_{n+1}(x) = \lambda_n(x) + \beta(n) \left(Q_n^{x}(x,1) - Q_n^{x}(x,0) \right).
\label{eq:learingW}
\end{equation}
Here, $s_n$ denotes the state visited at time step $n$, $x$ is a reference state for which we wish to learn Whittle's index, $r_0(s_n)$ and $r_1(s_n)$ are the sampled rewards for actions passive and active, respectively, and $\alpha(n)$ and $\beta(n)$ are learning parameters. The superscripts in (\ref{eq_q-learning with x}) stand for the parametric dependence of the Q-values on $x, \lambda$, where $x$ is a fixed reference state and $\lambda = \lambda(x)$ is a slowly varying parameter updated by \eqref{eq:learingW}. As usual in the literature, the learning parameters need to satisfy $\sum_n \alpha(n) = \infty$,  $\sum_n \alpha(n)^2 < \infty$, $\sum_n \beta(n) = \infty, \sum_n \beta(n)^2 < \infty$. In addition, we require 
 $\beta(n) = o(\alpha(n))$ in order to implement the two distinct time scales, namely, the relatively faster time scale for the updates of the state-action function, and the slow one for the Whittle indices. {In our case, we use the following stepsizes that meet the above-mentioned conditions:}

 \begin{align}
     \alpha(n) = & \frac{1}{\lceil{n/5000} \rceil} \label{eq:alpha learning rate}
     \\
     \beta(n) = & \frac{1}{1 + \lceil{n \log(n) / 5000}\rceil}I\{(n) \text{ mod}(50)=0\} \label{eq:beta learning rate}
 \end{align}

The pseudo-code implementation of QWI is present in Alg. \ref{alg:tabular QWI}. For a given value of the discount parameter $\gamma$ and exploration parameter $\epsilon$, we initialise the $N$ arms with a random state. At every decision epoch $n$, with probability $1-\epsilon$, we choose the greedy action, i.e, we activate ($a_n^i=1$) the $M$  arms  with largest $\lambda_n(x)$, while with probability $\epsilon$ we choose $M$ arms at random. Rewards $r_n^i$ are collected, the new states $s_{n+1}^i$ are observed, and the values of the state-action function and the Whittle indices are updated by (\ref{eq_q-learning with x}) and (\ref{eq:learingW}).
For a given arm $i$, updating the value functions of all states $s \in \mathcal{S}^i$ necessitates iterating over all states in the state space and utilizing all reference states $x$. Thus, the computational complexity, defined as the total number of computations required for these updates, scales as 
$O(|\mathcal{S}^i|^2\, |\mathcal{A}|)$, where $\mathcal{A}=\{0, 1\}$.
We note that QWI learns simultaneously the Whittle indices of all the states of every arm on-line.

\begin{algorithm}[H]
\caption{Tabular QWI Algorithm}\label{alg:tabular QWI}
\begin{algorithmic}
   \STATE {\bfseries Input:} Discount parameter $\gamma \in (0,1)$, exploration parameter $\epsilon \in [0,1]$, 
   \STATE {\bfseries Output:} Whittle index matrix for all states in each arm $i$
   \STATE Initialize $s_0$ for all arms
   \FOR{$n = 1:n_{end}$}
        \STATE Define action $a_n^i$ through $\epsilon$-greedy policy for each arm $i$
        \STATE Get new states $s_{n+1}^i$ and rewards $r_n^i$ from states $s_n^i$ and actions $a_n^i$
        \STATE Update learning rate $\alpha(n)$, $\beta(n)$ as (\ref{eq:alpha learning rate}) and (\ref{eq:beta learning rate})
        \STATE Update $(s_n^i, a_n^i)$ $Q$-values as (\ref{eq_q-learning with x})
        \STATE Update Whittle estimates for all states $x$ in each arm $i$ as (\ref{eq:learingW})
   \ENDFOR
   
\end{algorithmic}
\end{algorithm}

The next result shows that QWI converges to the Whittle index for any RMABP. The proof, adapted from \cite{Avrachenkov2022}, is given in Appendix \ref{sec:proof of convergence annex}.

\begin{theorem} (Convergence of QWI) Given learning parameters $\alpha(n)$ and $\beta(n)$ such that $\sum_n \alpha(n) = \sum_n \beta(n) = \infty$, $\sum_n \alpha(n)^2 < \infty$, $\sum_n \beta(n)^2 < \infty$ and $\beta(n) = o (a(n))$ and that the problem satisfies the indexability condition, iterations (\ref{eq_q-learning with x}) and (\ref{eq:learingW}) converge respectively  to the state-action function of Whittle index policy, denoted by $Q_W(s,a)$, and to the Whittle indices $\lambda(s)$, i.e., $\lambda_n(s) \rightarrow \lambda(s)$ and $Q_n(s,a) \rightarrow Q_W(s,a)$ a.s.\ $\forall s \in S, a \in A$ as $n \rightarrow \infty$.
\label{thmQWI}
\end{theorem}

\subsection{QWINN algorithm}\label{sec_QWINN}
The QWINN algorithm extends the Whittle index heuristic of the QWI algorithm through the integration of neural networks for approximating $Q$-values. Distinct from the tabular approach in section \ref{sec_tabularQWI}, QWINN (algorithm \ref{alg:QWINN}) utilizes a neural network with three hidden layers (100,200,100 neurons each) connected by ReLU activation functions. This network inputs two state variables, the visited state $s$ and the reference state $x$ and outputs $Q$-values for both possible actions $Q_\theta^x(s) = \left[Q_\theta^x(s,0) \phantom{-} Q_\theta^x(s,1)  \right]$. Following QWI's heuristic, QWINN updates its Q-value neural network estimator as:
\begin{equation}\label{eq_target NNQvalue}
    Q_{target}^x(s_n,a_n) = (1-a_n)(r_0(s_n)+ \lambda(x)) + a r_1(s_n) + \gamma \max_{v \in A} Q_{\theta'}^x(s_{n+1}, v).
\end{equation}
QWINN utilizes a tabular approach to update the Whittle indices, similar to QWI's heuristic in Equation \eqref{eq:learingW}. The Whittle indices are defined as:
\begin{equation}\label{eq:whittle update qwinn}
    \lambda_{n+1}(x) = \lambda_n(x) + \beta(n) \left( Q_\theta^{x}(x,1) - Q_\theta^{x}(x,0) \right).
\end{equation}

One of the main changes with respect to the tabular algorithm in section \ref{sec_tabularQWI} is the use of Double Q-Learning \cite{van2016deep} \cite{Hasselt2010}, where we employ a second neural network for the computation of the $\max_{v \in A}Q_{\theta'}^x(s_{n+1}, v)$ term in Equation~(\ref{eq_target NNQvalue}). {The reason for this decision is due to the maximization bias of Q-learning: overestimating the Q-values results in this error increase over time, due to the target being $r + \gamma \max_v Q^x(s,v)$. The use of a second neural network for the target helps control this bias}. This second neural network copies the parameter values of the main network $Q^x_{\theta}$ every 50 iterations.

{
In the following analysis, we investigate the local convergence properties of the DQN algorithm. We strengthen a necessary condition for local minima in optimization theory to an assumption and apply stochastic approximation theory to elucidate DQN's behavior in a neighborhood of a local minimum under specific conditions.
\subsubsection{Preliminary Considerations}

Consider the sequence $\tilde{\theta}_m = \theta_{T_n}$, defined for each $T_n \leq m < T_{n+1}$, where $T_n \uparrow \infty$. We specify that for some $n$, the parameter $\theta^*_n := \theta_{T_n}$ lies in a bounded neighborhood around a local minimum $\theta^*$ of the Bellman error function $\mathcal E(\cdot, \theta^*)$, defined as:
$$
\mathcal E(\theta,\tau) := \mathbb E\left[\left\|Q^{x}_\theta(s,a) - (1-a)(r_0(s)+\lambda) - a r_1(s) - \gamma\max_{v\in A}Q^{x}_\tau(s',v)\right\|^2\right].
$$
Let $\nabla_1\E, \nabla_1^2\E$ denote the gradient and the Hessian of $\E$ with respect to the first argument alone. Assuming that the Hessian $\nabla^2_1 \mathcal E(\theta^*, \theta^*)$ is positive definite, the inverse function theorem ensures a locally defined, bijective mapping $F(\tau) = (\nabla_1\E(\cdot,\tau))^{-1}(\textbf{0})$, where $\textbf{0}$ is the zero vector, in a neighborhood of $\theta^*$. Let us introduce an additional assumption.

{ 
\begin{asum} 
\label{asum_contr}
$F(\tau)$ is locally a contraction around the equilibrium point $(\theta^*,\theta^*)$.
\end{asum}

}


\subsubsection{Local Convergence}
\begin{theorem}
\label{thm:conv}
    {For a $\theta^*$ as above, under Assumption~\ref{asum_contr} and the condition that the Hessian $\nabla^2_1 \mathcal E(\theta^*, \theta^*)$ is positive definite, {the DQN algorithm exhibits local convergence to an open ball of radius $\frac{2\epsilon}{1-\alpha}$ centered at  $\theta^*$, for $T_{n+1}-T_n$ sufficiently large.}}
\end{theorem}

\begin{proof} {By Assumption \ref{asum_contr}, $F(\theta^*)$ is locally a contraction with some factor}  $0 < \alpha < 1$, and assuming that $\epsilon$ is within the specified bounds, consider $F_n(\theta^*_n) := (\nabla_1 \mathcal E(\cdot, \theta^*_n))^{-1}(\textbf 0)$. For $T_{n+1} - T_n$ sufficiently large, $\|\theta^*_{n+1} - F_n(\theta^*_n)\| < \epsilon$. {  By continuity, for $\theta$ in the $\epsilon$-neighbourhood of $\theta^*$, $\|F_n(\theta) - F_n(\theta)\| < \epsilon'$, for some $\epsilon'$ that we take to equal $\epsilon$ without loss of generality. } Then
\begin{eqnarray*}
\|\theta^*_{n+1} - \theta^*\| &\leq& \epsilon + \|F_n(\theta^*_n) - F(\theta^*)\| \\
&\leq& \epsilon + \|F_n(\theta^*_n) - F(\theta_n^*)\| + \|F(\theta_n^*) - F(\theta^*)\| \\
&\leq& 2\epsilon + \alpha\|\theta^*_n - \theta^*\|.
\end{eqnarray*}
Iterating, it follows that $\theta^*_n$ approaches the $\delta$-ball centred at $\theta^*$. This establishes the local convergence of the DQN algorithm within a neighborhood $B$ of $\theta^*$ to a $\delta$-neighborhood thereof. \end{proof} 

\subsubsection{Observations and Practical Implications}

{One of the key assumptions we have introduced in Theorem~\ref{thm:conv} is that $ F(\tau) $, defined as $ F(\tau) = (\nabla_1 \mathcal E(\cdot,\tau))^{-1}(\textbf{0}) $, is locally a contraction around the point $ \theta^* $. This assumption is supported by numerical analysis, which shows that in our examples, $ F(\tau) $ is indeed a contraction around $ \theta^* $. However, finding sufficient general conditions for this contraction to hold falls outside the scope of this paper. 

We have performed numerical calculations to illustrate that this assumption holds in the problems considered in the paper. In order to do so, we have proceed as follows. Linearizing $\nabla_1 \mathcal E(\theta, \tau)$ in the proximity $(\theta^*,\theta^*)$ we get
$$
\nabla_1 \mathcal E(\theta, \tau) \approx \nabla_1 \mathcal E(\theta^*, \theta^*) + \nabla_1^2 \mathcal E(\theta^*, \theta^*) (\theta-\theta^*) + \nabla_2\nabla_1 \mathcal E(\theta^*, \theta^*) (\tau-\theta^*) + o(\theta-\theta^*,\tau-\theta^*  ).
$$
Since, $\nabla_1 \mathcal E(\theta^*, \theta^*)=0$, we get
$$
(\theta-\theta^*) \approx - (\nabla_1^2 \mathcal E(\theta^*, \theta^*))^{-1} \nabla_2\nabla_1 \mathcal E(\theta^*, \theta^*) (\tau-\theta^*).
$$
To have a contraction (locally) around the equilibrium point, we need the spectrum of $- (\nabla_1^2 \mathcal E(\theta^*, \theta^*))^{-1} \nabla_2\nabla_1 \mathcal E(\theta^*, \theta^*)$  to be inside the unit disc in the complex plane around the origin.

The foregoing assumes that $\mathcal{E}$ is differentiable in the second argument, i.e., the target. More generally, a version of Danskin's theorem ensures existence of directional derivatives and the foregoing can be modified accordingly.

To calculate numerically the eigenvalues we have considered a random batch of training samples and compute the loss function. 
{We then compute numerically the gradient of the loss function with respect to  the first argument  ($\nabla_1 \mathcal E(\cdot, \cdot)$) and then again, the gradient thereof with respect to  the first argument ($\nabla_1^2 \mathcal E(\cdot, \cdot)$) and the second argument ($\nabla_2\nabla_1 \mathcal E(\cdot, \cdot)$).}


We have numerically verified that the condition on the spectrum for all the environments considered in the paper is satisfied. 
In the following histogram, we represent the eigenvalues obtained using the homogeneous environments in our paper, using a neural network with size $((2, 50), (50,100), (100, 50), (50,2))$, that is, with 10402 parameters.

Furthermore, the practical application of this contraction depends on the intervals $ T_{n+1} - T_n $ being sufficiently large. The larger these intervals are, the smaller the choice of $ \delta $ can be. This observation may not be consistent with settings where $ T_n = nT $ is used for some fixed $ T > 0 $ with decreasing step sizes. Nevertheless, it remains valid when a constant step size is employed, provided that $ T $ is sufficiently large.}

\begin{figure}[ht]
    \centering
    \begin{subfigure}[b]{0.32\textwidth}
        \centering
         \includegraphics[width=\textwidth]{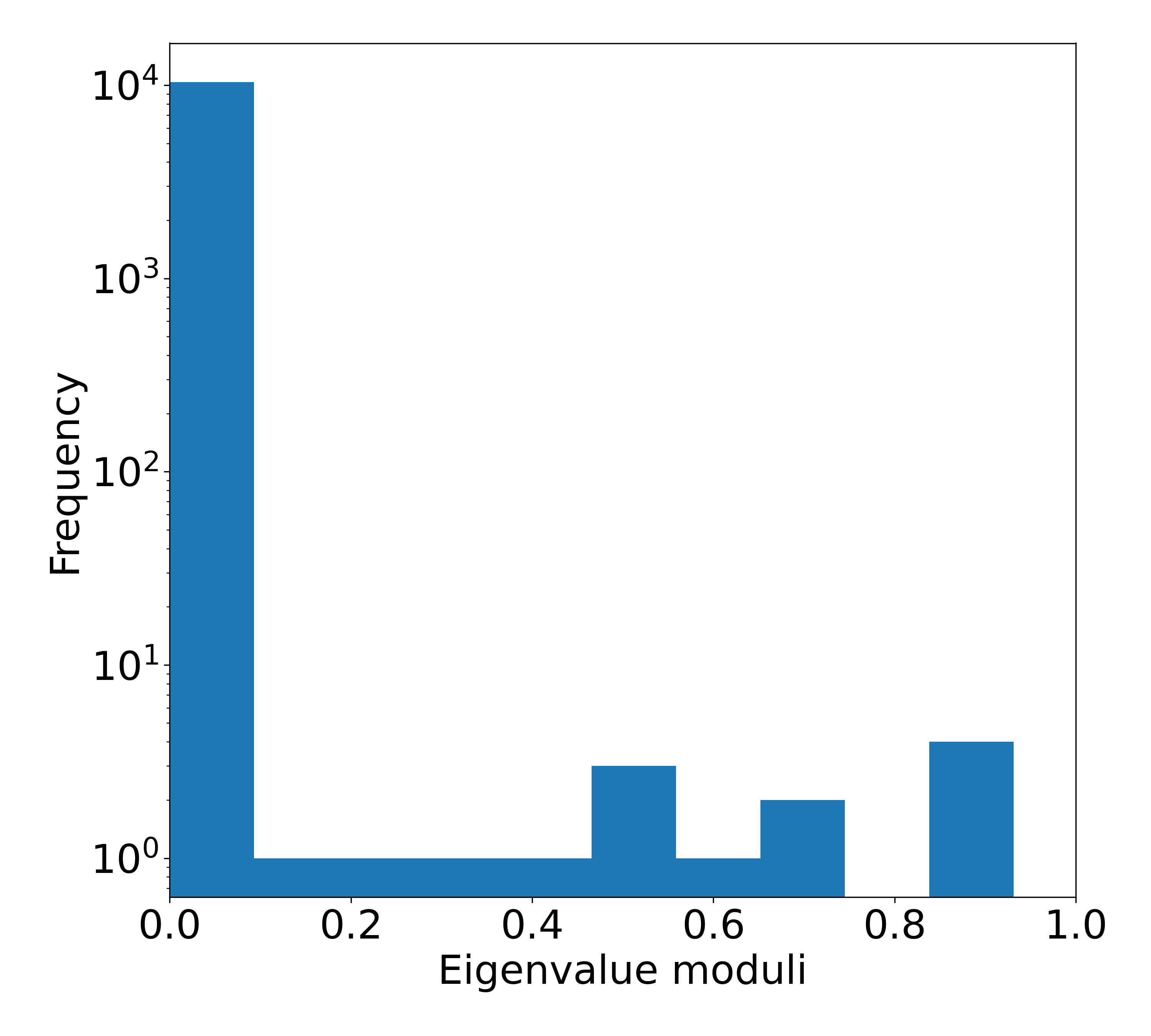}
        \caption{Circular problem}
        \label{fig:eigenvalue_circular}
    \end{subfigure}
    \hfill 
    \begin{subfigure}[b]{0.32\textwidth}
        \centering
        \includegraphics[width=\textwidth]{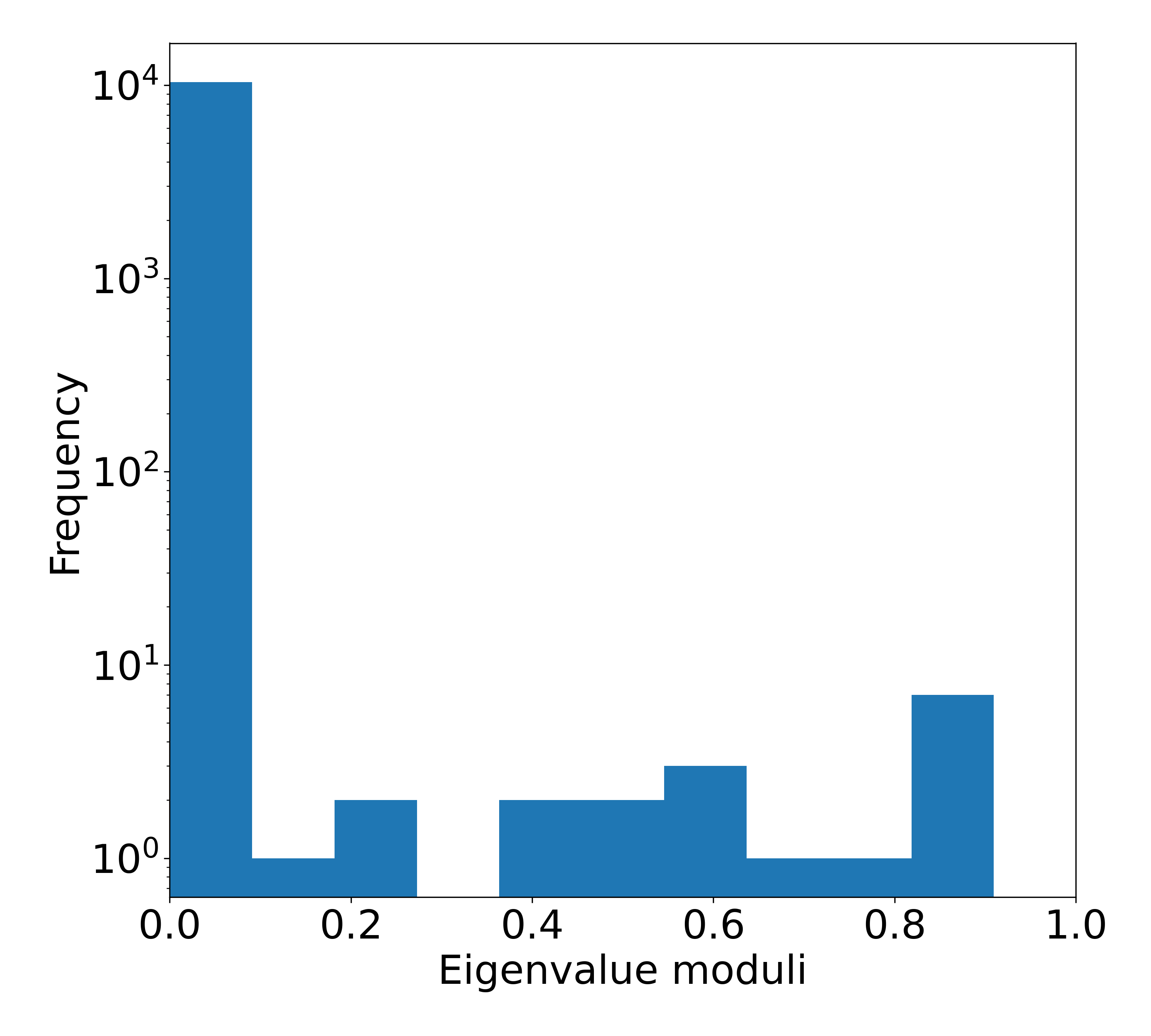}
        \caption{Restart problem}
        \label{fig:eigenvalue_restart}
    \end{subfigure}
    \hfill 
    \begin{subfigure}[b]{0.32\textwidth}
        \centering
        \includegraphics[width=\textwidth]{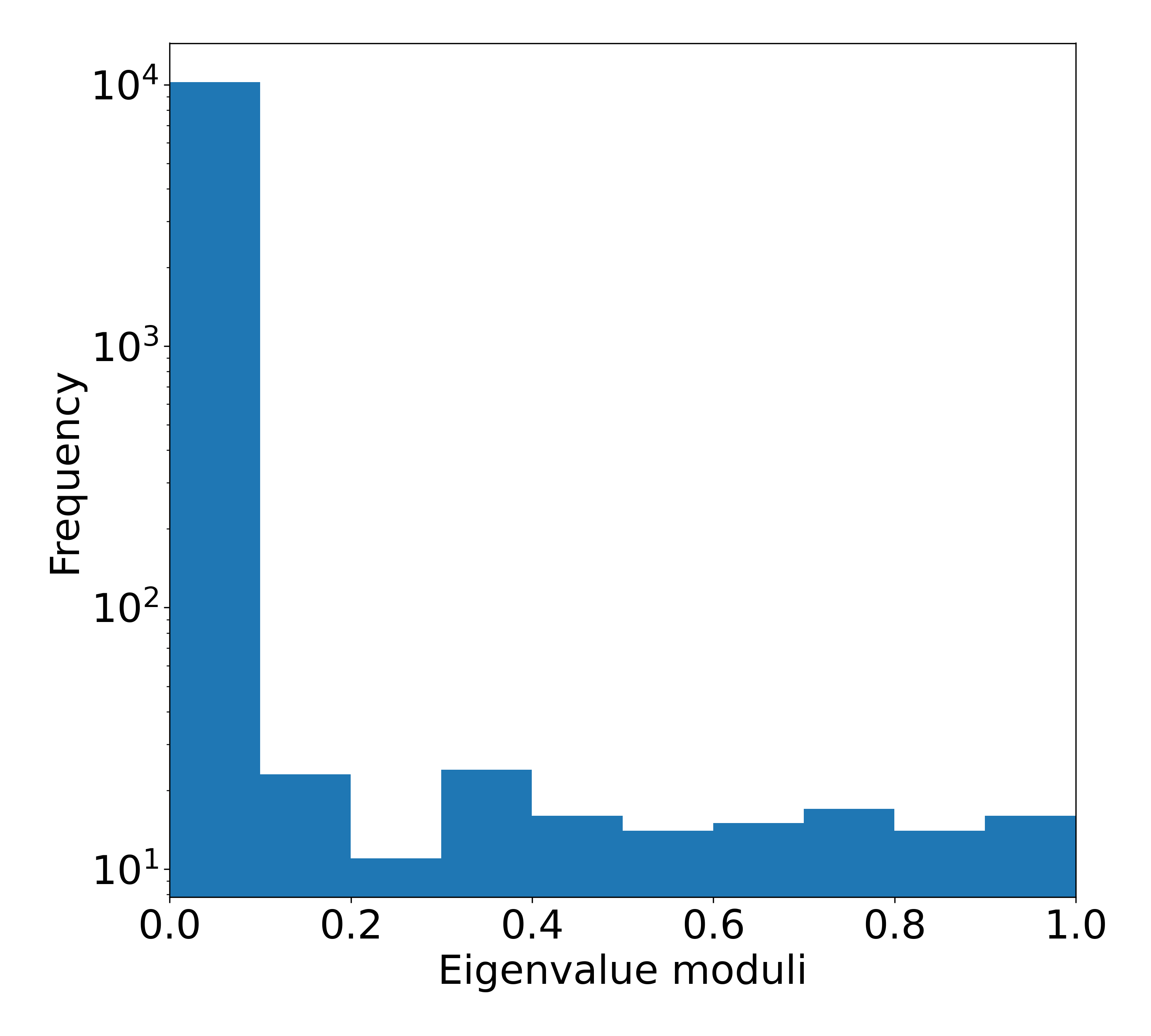}
        \caption{Deadline Scheduling problem}
        \label{fig:eigenvalue_deadline}
    \end{subfigure}
    \caption{Histogram of eigenvalue moduli of $-(\nabla_1^2 \mathcal E(\theta^*, \theta^*))^{-1} \nabla_2\nabla_1 \mathcal E(\theta^*, \theta^*)$}
    \label{fig:eigenvalues}
\end{figure}

}

\begin{algorithm}[h]
\caption{QWINN Algorithm}\label{alg:QWINN}
\begin{algorithmic}
   \STATE {\bfseries Input:} Discount parameter $\gamma \in (0,1)$, exploration parameter $\epsilon \in [0,1]$, 
   \STATE {\bfseries Output:} Whittle index vector for all states
   \STATE Initialize $s_0$ for all arms
   \FOR{$n=1:n_{end}$}
        \STATE Define action $a_n^i$ through $\epsilon$-greedy policy for each arm $i$
        \STATE Get new states $s_{n+1}^i$ and rewards $r_n^i$ from states $s_n^i$ and actions $a_n^i$
        \STATE Save each arm's data into separated memories
        \IF {number of samples in memory $>$ threshold}
            \STATE Update Whittle index learning rate $\beta(n)$ as (\ref{eq:beta learning rate})
            \FOR{arm $i \in N$}
                \FOR{$x \in S$}
                    \STATE Predict $Q$-values of sample batch $(s_n^i, a_n^i)$ using reference state $x$ for Whittle index with $Q_x^x$
                    \STATE Compute target $Q$-value for sample $(s_n^i, a_n^i, r_n^i, s_{n+1}^i, x)$ as (\ref{eq_target NNQvalue}), using secondary network $Q_{\theta'}^x$
                \ENDFOR
                \STATE Compute {the mean square error} loss function between $Q_\theta$ and $Q_{target}$
                \STATE Update the $\theta$ parameters of the $Q_\theta$ regressor through backpropagation {using an Adam optimizer}
                \STATE Update the Whittle index for all states $x$ as (\ref{eq:whittle update qwinn})
                \IF {$n \% 50 = 0$}
                    \STATE Copy the $\theta$ parameters from the main $Q_\theta^x$ neural network into the secondary $Q_{\theta'}^x$ neural network
                \ENDIF
            \ENDFOR
        \ENDIF
   \ENDFOR
   
\end{algorithmic}
\end{algorithm}

\subsection{Other algorithms}\label{sec_other algorithms}
We describe now the other algorithms from the literature that we will consider in the numerical simulations.
\subsubsection{NeurWIN algorithm}\label{sec_NeurWIN algorithm}
The NeurWIN algorithm, introduced in \cite{Nakhleh2021}, is a neural network-based algorithm designed to directly compute the Whittle indices' estimates $\lambda_\theta(s_n)$ of a problem using as an input to the neural network only the state whose index is to be computed. 

The NeurWIN algorithm considers that a policy of indices that achieves optimal rewards for RMABP is equivalent to Whittle's index policy. To obtain this policy, they define an activation function using a sigmoid function:

\begin{equation}
\sigma_m(f_\theta(s[t]) - \lambda) = \frac{1}{1 + e^{-m(f_\theta(s[t]) - \lambda)}}, \label{eq: activation sigma neurwin}
\end{equation}
where $m$ is a sensitivity parameter. This function selects action $a=1$ with probability $\sigma_m(f_\theta(s[t]) - \lambda)$ and $a=0$ with probability $1-\sigma_m(f_\theta(s[t]) - \lambda)$. For each mini-batch of episodes, they randomly choose two states $s_0$ and $s_1$, with $s_0$ serving as the fixed value $\lambda=f_\theta(s_0)$ and $s_1$ as the initial state. Multiple episodes are generated within each mini-batch, recording the actions and states visited based on the policy defined by Eq. \eqref{eq: activation sigma neurwin}. The gradients $h_e$ are calculated as follows:

\begin{equation}
h_e \leftarrow \begin{cases}
h_e + \nabla_\theta \ln{(\sigma_m(f_\theta(s[t])-\lambda))} & \text{if } a[t] = 1 \\
h_e + \nabla_\theta \ln{(1 - \sigma_m(f_\theta(s[t])-\lambda))} & \text{if } a[t] = 0
\end{cases}
\end{equation}
for each sample at time $t$ within the episode. After all episodes in the mini-batch are completed, the neural network parameters are updated as follows:

\begin{equation}
\theta \leftarrow \theta + L_b \sum_e(G_e - \bar G_b)h_e,
\end{equation}
where $G_e$ represents the discounted net rewards, $G_b$ is the average of these rewards, and $L_b$ is the learning rate.

In order to compare the convergence speed of all these algorithms to the desired policy, we have decided to represent each NeurWIN update as a unique transition, so that at each iteration all algorithms are updated simultaneously. It is also important to highlight that, unlike in QWI and QWINN, the training and execution phases are separate in NeurWIN, i.e., we first need to learn the indices of each arm separately.

\subsubsection{$Q$-learning and DQN algorithms}\label{sec_DQN algorithm}
We also consider the vanilla implementations of $Q$-learning and DQN \cite{sutton2018reinforcement}. $Q$-learning is 
known to converge to the optimal policy,  \cite{watkins1992q}, but it suffers from the ``curse of dimensionality''. DQN does not have performance guarantees, but in practice performs better thanks to the interpolation of the state information.

As with QWINN, DQN training is performed through batches of samples stored in memory, where the input to the neural network is the combination of arm states, while the outputs are the $Q$-values for each possible combination of arm activations. In the following results, $Q$-learning uses 
$\alpha = \frac{1}{\# N(s,a)}$ 
as the learning step size, where 
$\# N(s,a)$ 
is the number of times that said state $(s,a)$ has been visited, while DQN has been trained using a fixed learning rate $lr=0.001$.


\section{Numerical results}\label{sec_numericalresults}
In this section we compare the performance of our algorithms QWI and QWINN with respect to $Q$-learning, DQN and NeurWIN. We consider three different RMABPs used in previous literature: the "restart problem" proposed in \cite{Avrachenkov2022}, the "deadline scheduling problem" studied in \cite{yu2018deadline} and the ``circular environment problem'' in \cite{fu2019towards}. 
An interesting feature of these examples is that the Whittle index can be calculated in closed-form, which allows us to assess the accuracy of the estimates of the Whittle indices obtained with the various algorithms.

We recall that Whittle index policy is a heuristic, i.e., it does not characterize the optimal solution to an RMABP, even though it has often been reported that its sub-optimality gap is very small. For this reason, we consider as a baseline the performance of the optimal policy $\pi^*$ wherever the latter can be computed. We recall that the optimal policy $\pi^*$ solves Bellman's Optimality Equation, see for example \cite{Puterman2014},

\begin{equation}\label{eq:value function optimal}
\begin{aligned}
    V_{\pi^*}(s) =& \max_{a \in \{0,1\}^N} \left(r(s,a) + \gamma \sum_{s'} p(s'|s,a) V_{\pi^*}(s') \right),
    \\
    &\text{s.t. } \sum_{i=1}^{N} a^i(t) = m, \forall t,
\end{aligned}
\end{equation}
where $s=(s^1,\dots,s^N)$ denotes the states of all the arms, $r(s,a)=\sum_{i=1}^N r^i(s^i,a^i)$, and $p(s'|s,a)= \prod_{i=1}^N p^i(s_i'|s_i,a_i)$. We solve equation~(\ref{eq:value function optimal}) by value Iteration.

We also use Bellman's equation in order to assess the performance of each algorithm during its training for small enough problems. Let $\pi$ denote the current policy of each one of the algorithms at a given time. Then, the value function $V_\pi(s)$ characterizing its performance is the solution of Bellman's equation
\begin{equation}\label{eq:value function policy}
    V_\pi(s) = \sum_a \pi(a|s) \left(r(s,a) + \gamma \sum_{s'} p(s'|s,a) V_\pi(s') \right),
\end{equation}
where $\pi(a|s)$ is the probability of taking action $a \in  \{0,1\}^N$ in state $s$. We also solve Equation (\ref{eq:value function policy}) using value Iteration.

In order to compare the performance of a policy $\pi$ with respect to the optimal policy $\pi^*$ we use the Bellman Relative  Error defined as
\begin{equation}
    BRE(\pi,\pi^*) = \frac{1}{|\mathcal{S}|} \sum_{s \in \mathcal{S}} \frac{ |V_\pi(s) - V_{\pi^*}(s) | }{V_{\pi^*}(s)}.
    \label{eq:relbelerror}
\end{equation}
{This metric will be particularly useful in subsections \ref{sec_homogeneous restart}, \ref{sec_heterogeneous restart} and \ref{sec:circular environment}, where the reduced state space allows us to easily compute it, and to study the  convergence of QWI, QWINN, NeurWIN, Q-learning and DQN  to the optimal policy.} 
Throughout this section, we will denote by $\pi_n^P$, $P \in \{\textrm{QWI}, \textrm{QWINN}, \textrm{DQN}, \textrm{Q} \}$, the Whittle index heuristic estimated by algorithm $P$ at time $n$.
We consider a discount factor $\gamma=0.9$ and an exploration parameter $\epsilon=1$ (i.e., pure off-policy mode).

\subsection{Restart problem (homogeneous arms)}\label{sec_homogeneous restart}
In this section we consider a ``restart problem'' with state space  $S=\{0,1,2,3,4\}$. 
From any state, the active action  ($a=1$) brings the arm to the initial state with probability $1$. i.e., $p(0|s,1)=1$, for all $s$. The transitions probability matrix in the case of passive action is:
\[
	P_0=\begin{pmatrix}
			1-x & x & 0 & 0 & 0\\
			1-x & 0 & x & 0 & 0\\
			1-x & 0 & 0 & x & 0\\
			1-x & 0 & 0 & 0 & x\\
			1-x & 0 & 0 & 0 & x
		\end{pmatrix}.
\]

The expected conditional reward is given by $r(s,0)=y^{s+1}$ for passive actions, while $r(s,1)=0$ for active actions ($a=1$). 
We consider the following setting: $N=5$ arms; one arm $M=1$ is activated every time epoch; and $x=y=0.9$. The Whittle index can be analytically calculated in this problem, yielding $\lambda(0)=-0.9$, $\lambda(1)=-0.7371$, $\lambda(2)=-0.5373$, $\lambda(3)=-0.3188$ and $\lambda(4)=-0.0939$ while using a discount factor $\gamma=0.9$. We note that the Whittle index is increasing as the state increases, which is consistent with the reward structure. We note that from the description of the problem, a good policy will tend to keep all arms around state 0, as the rewards are larger there.

\begin{figure}[h]
    \centering
    \begin{subfigure}{0.49\textwidth}
        \includegraphics[width=\textwidth]{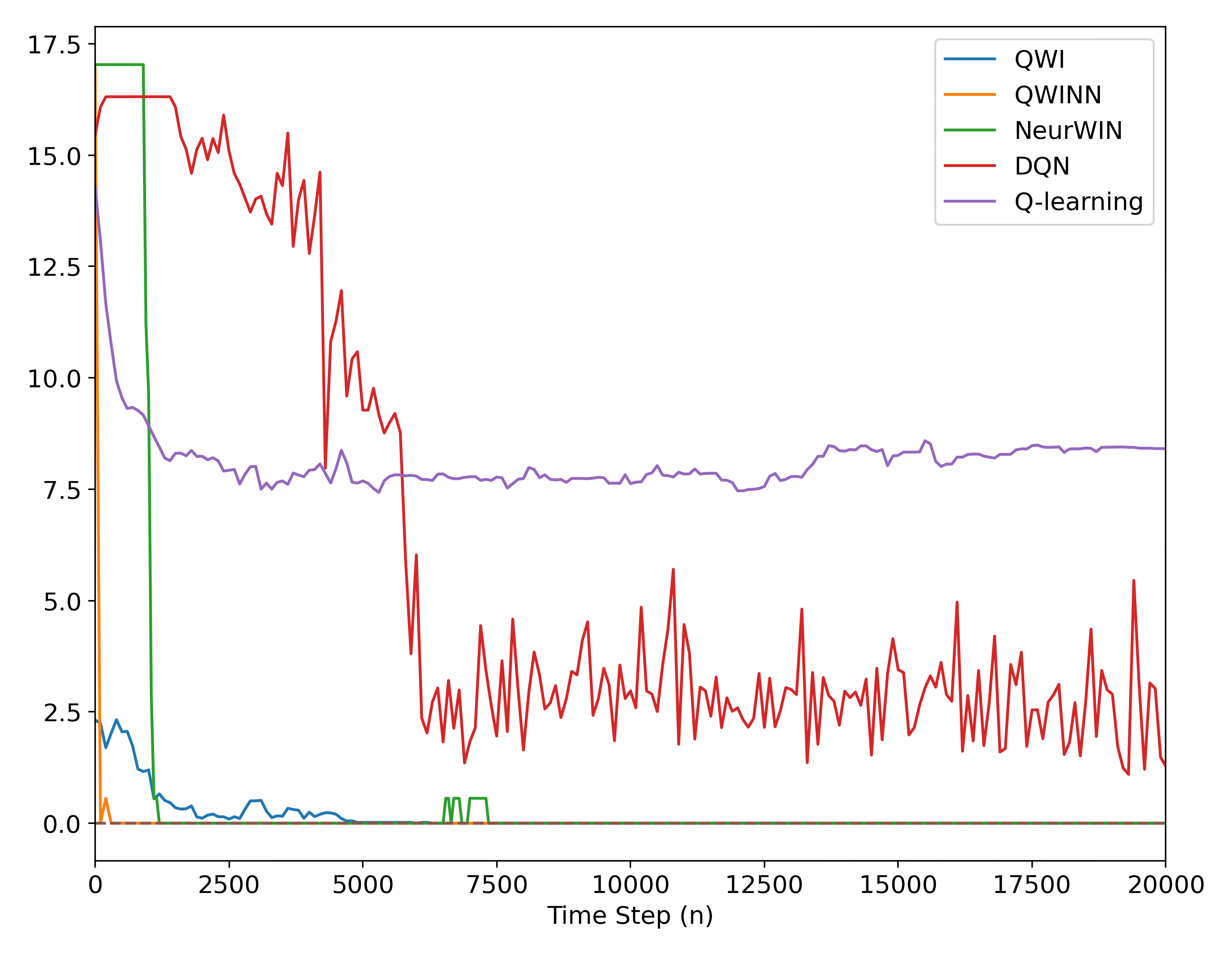}
    \end{subfigure}
    \hfill
    \begin{subfigure}{0.49\textwidth}
        \includegraphics[width=\textwidth]{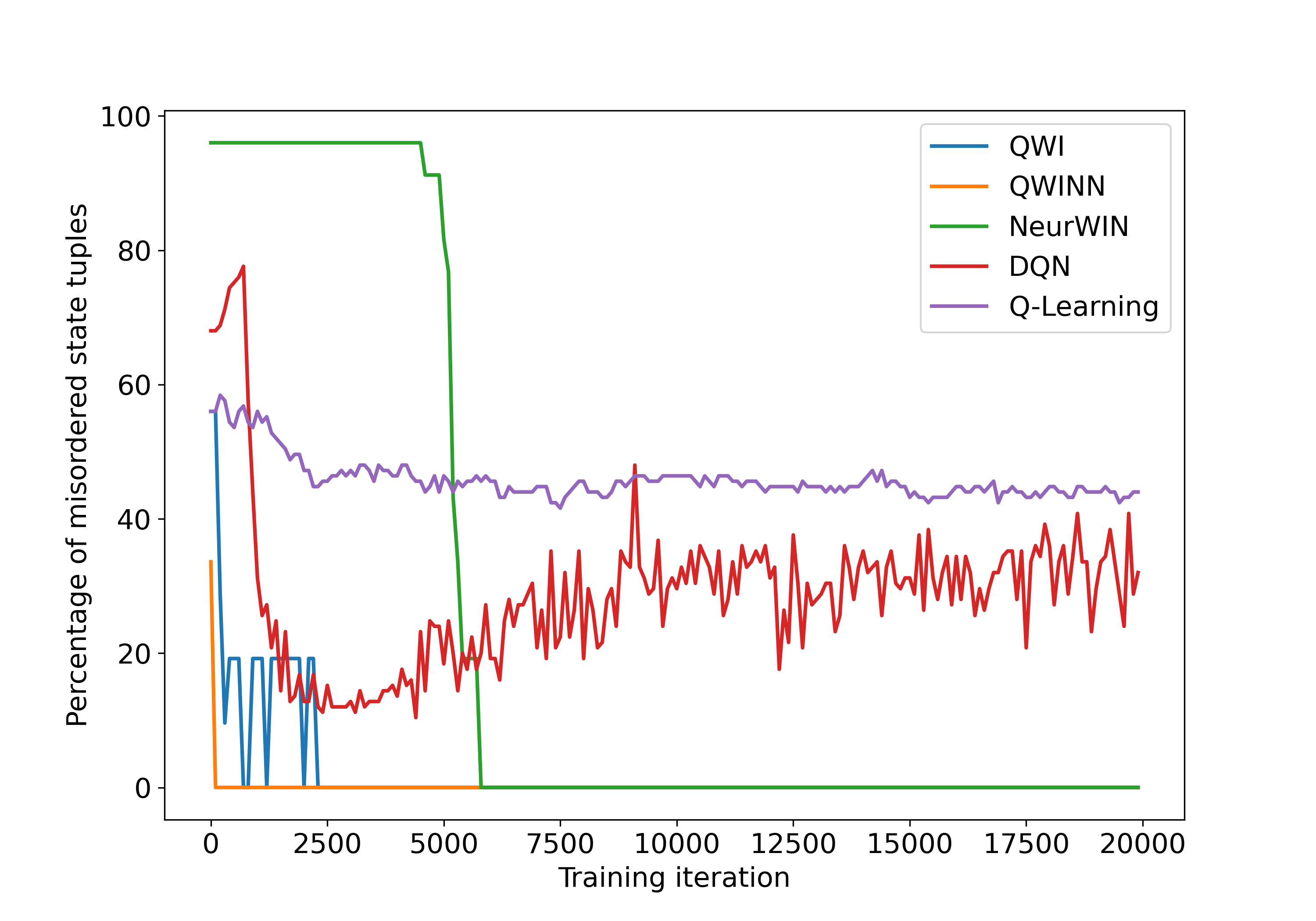}
    \end{subfigure}
        \caption{Performance graphs for the ``homogeneous restart'' problem. \emph{(left)} Bellman Relative Error $BRE(\pi_n^P)$, $P\in \{\textrm{QWI}, \textrm{QWINN}, \textrm{NeurWIN}, \textrm{DQN}, \textrm{Q-learning}  \}$ during training for the ``homogeneous restart'' problem, $N=5, M=1, |S|=5$; and \emph{(right)} Percentage of states in which an optimal action is not performed in the ``restart'' problem for homogeneous arms  }
        \label{fig:homogeneous restart performance}
\end{figure}

\begin{figure}[h]
    \centering
    \begin{subfigure}{0.49\textwidth}
        \includegraphics[width=\textwidth]{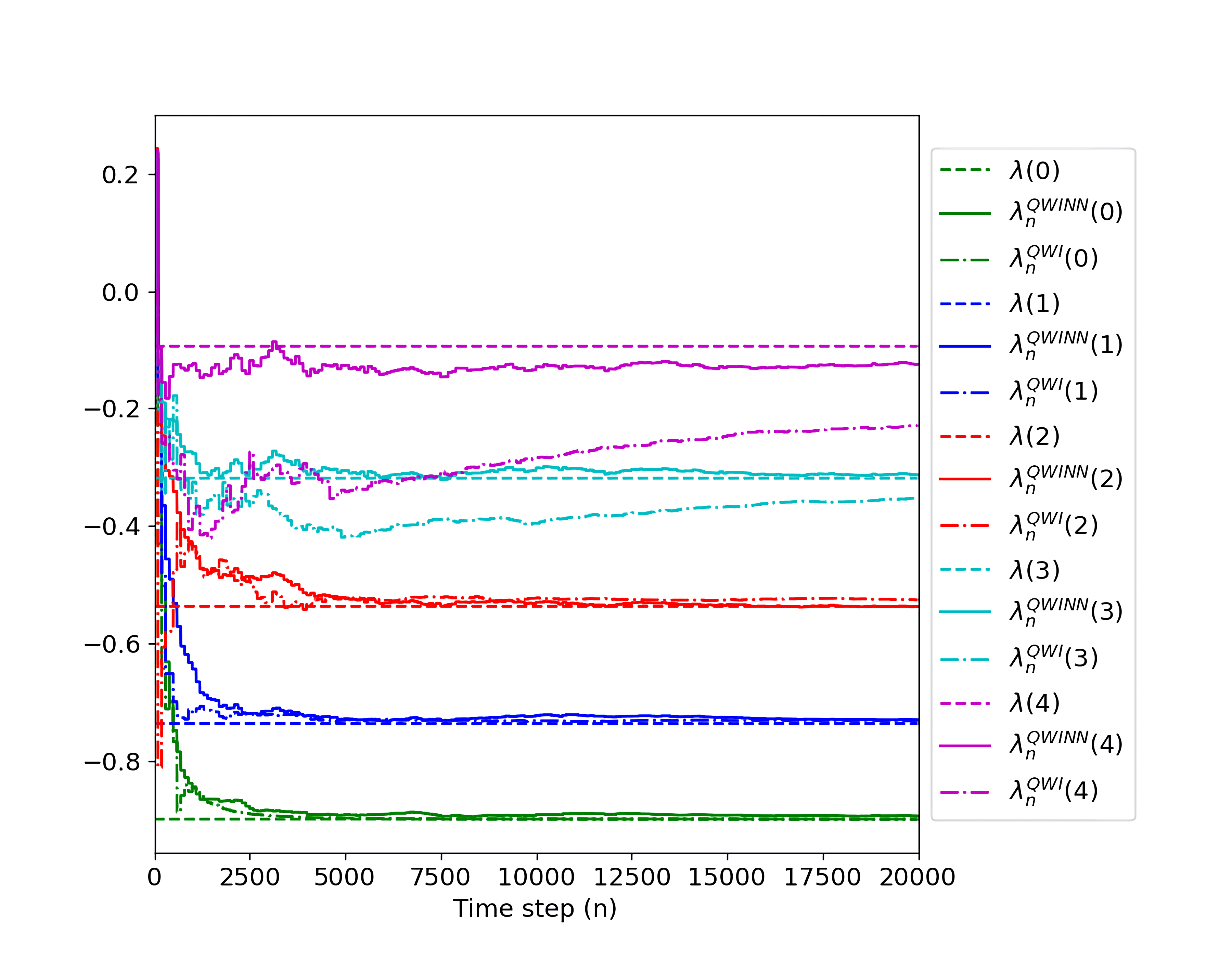}
        \caption{QWINN (solid lines) and QWI (dashdot lines)}
        \label{fig_restart index qwi - sub1}
    \end{subfigure}
    \hfill
    \begin{subfigure}{0.49\textwidth}
        \includegraphics[width=\textwidth]{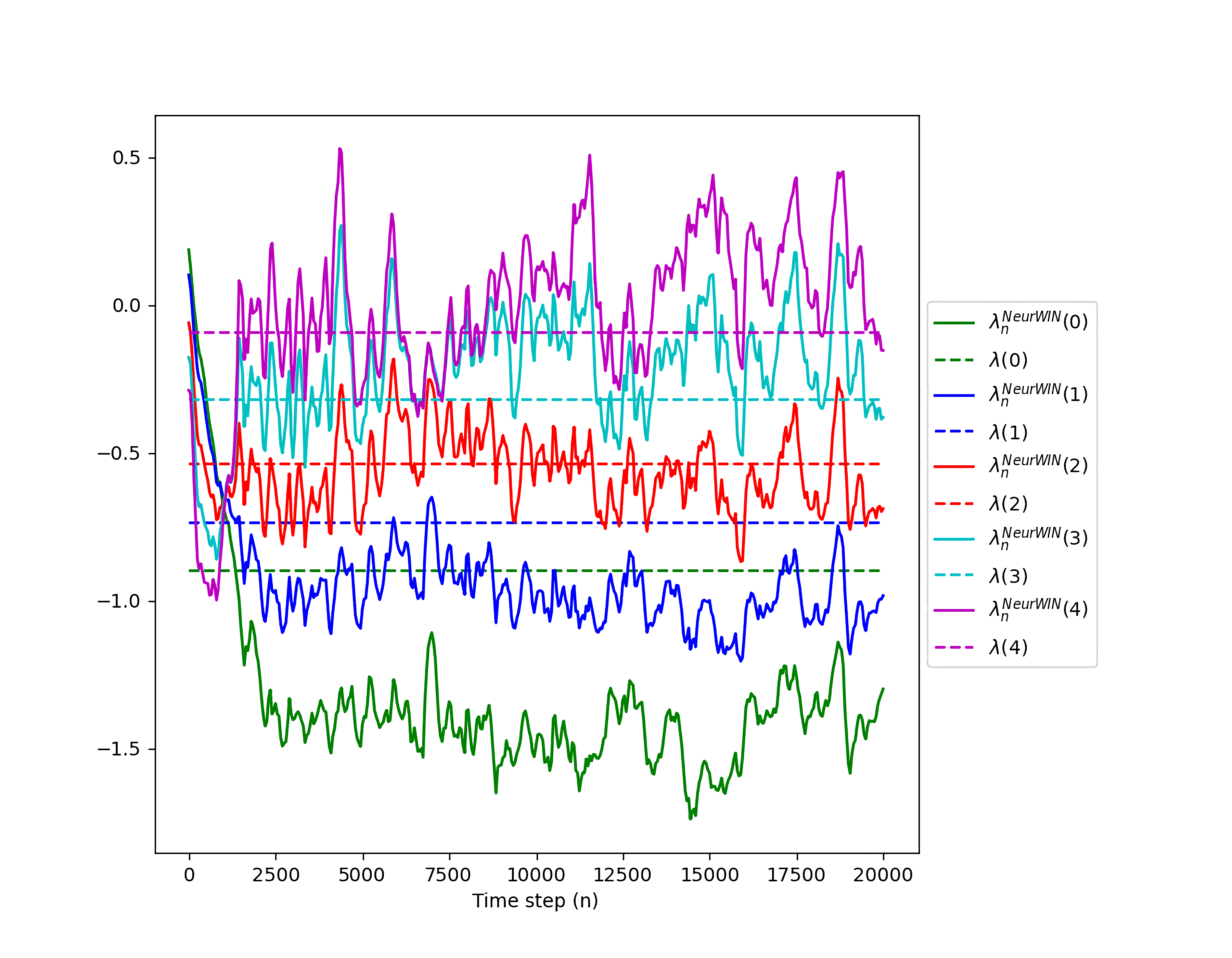}
        \caption{NeurWIN}
        \label{fig_restart index neurwin - sub1}
    \end{subfigure}
    \caption{Evolution of the Whittle index estimates for the restart problem}
\end{figure}

In Figure~\ref{fig:homogeneous restart performance} \emph{(left)} we depict the Bellman Relative Error for all the algorithms, namely, we depict $BRE(\pi_n^P,\pi^*)$, $P\in \{\textrm{QWI}, \textrm{QWINN}, \textrm{NeurWIN}, \textrm{Q-learning}, \textrm{DQN}  \}$, for all the time steps $n$, {while Figure \ref{fig:homogeneous restart performance} \emph{(right)} represents the percentage of states in which an optimal action is not taken}. We observe that QWI, NeurWIN and QWINN learn the optimal policy, QWINN being the fastest. We note that even though Whittle index policy need not be optimal, it might happen in such a  simple problem. On the other hand, $Q$-learning does not learn the optimal policy, even though we know from theory that as $n\to \infty$, it is the only algorithm among the approaches considered here for which convergence to the optimal policy is guaranteed \cite{watkins1992q}.  
The vanilla DQN implementation does better, but it requires longer to converge to the optimal policy. 
 
An important observation is that the performance of an index policy, in particular the values plotted in Figure~\ref{fig:homogeneous restart performance}~\emph{(left)}, only depend on the relative ordering between the states, and not on the precise value of the Whittle index. 
We thus explore the accuracy of the estimates  of QWI/QWINN (see Figure~\ref{fig_restart index qwi - sub1}) and NeurWIN (see Figure~\ref{fig_restart index neurwin - sub1}) during their training. 
More specifically, in Figure~\ref{fig_restart index qwi - sub1} we plot the averaged  (across the 5 arms) estimates of Whittle indices.
In the case of NeurWIN, there is a single agent cloned for the 5 arms, and therefore, in Figure~\ref{fig_restart index neurwin - sub1} we depict the estimate of this single agent.

In Figure~\ref{fig_restart index qwi - sub1} we note that the QWI's estimates for the indices of states 0,1 and 2 converge, whereas the estimates of states 3 and 4 need longer simulations due to the fact that these states are visited less frequently, while QWINN converges rapidly to the theoretical values of all the indices. It is not surprising that the estimates for states 3 and 4 require longer as they are visited less frequently.
We also observe that the estimates are stable, i.e, there are no fluctuations. In Figure~\ref{fig_restart index neurwin - sub1} we note that the ordering with NeurWIN is correct, but that the fluctuations do not vanish. The latter might suggest that  in the case of heterogeneous arms, NeurWIN might swap the ordering of the states, which might yield a sub optimallity gap larger than with QWI and QWINN. In Section~\ref{sec_heterogeneous restart} we show that this is indeed the case.

\subsection{Restart problem (heterogeneous arms)}\label{sec_heterogeneous restart}
In this section we want to investigate the impact that errors in the estimation of the Whittle index has on the performance of the arms. In order to do so, we consider a restart problem with $N=3$ arms, each one having a different set of parameters, and as before we consider $M=1$. As in the previous section, we assume that for every arm $x=y$, and that this value is taken uniformly at random from the set $\{0.4, 0.5, 0.6, 0.7, 0.8, 0.9\}$. We simulate 20 such instances, and to assess the sub-optimality gap we consider the Bellman Averaged Relative Error of algorithm $P$ defined as:
\begin{equation}
    BRE(P) = \frac{1}{20}\sum_{k=1}^{20} BRE_k(\pi_k^P,\pi_k^*),
    \label{eq:avgrelbelerror}
\end{equation}
where $BRE_k(\pi_k^P,\pi_k^*)$ denotes the Bellman Relative Error for instance $k$, $\pi_k^*$ is the optimal policy for instance $k$, and $\pi_k^P$ is the Whittle Index policy for instance $k$ obtained with algorithm $P$.





\begin{figure}[h]
    \centering
    \begin{subfigure}{0.49\textwidth}
        \includegraphics[width=\textwidth]{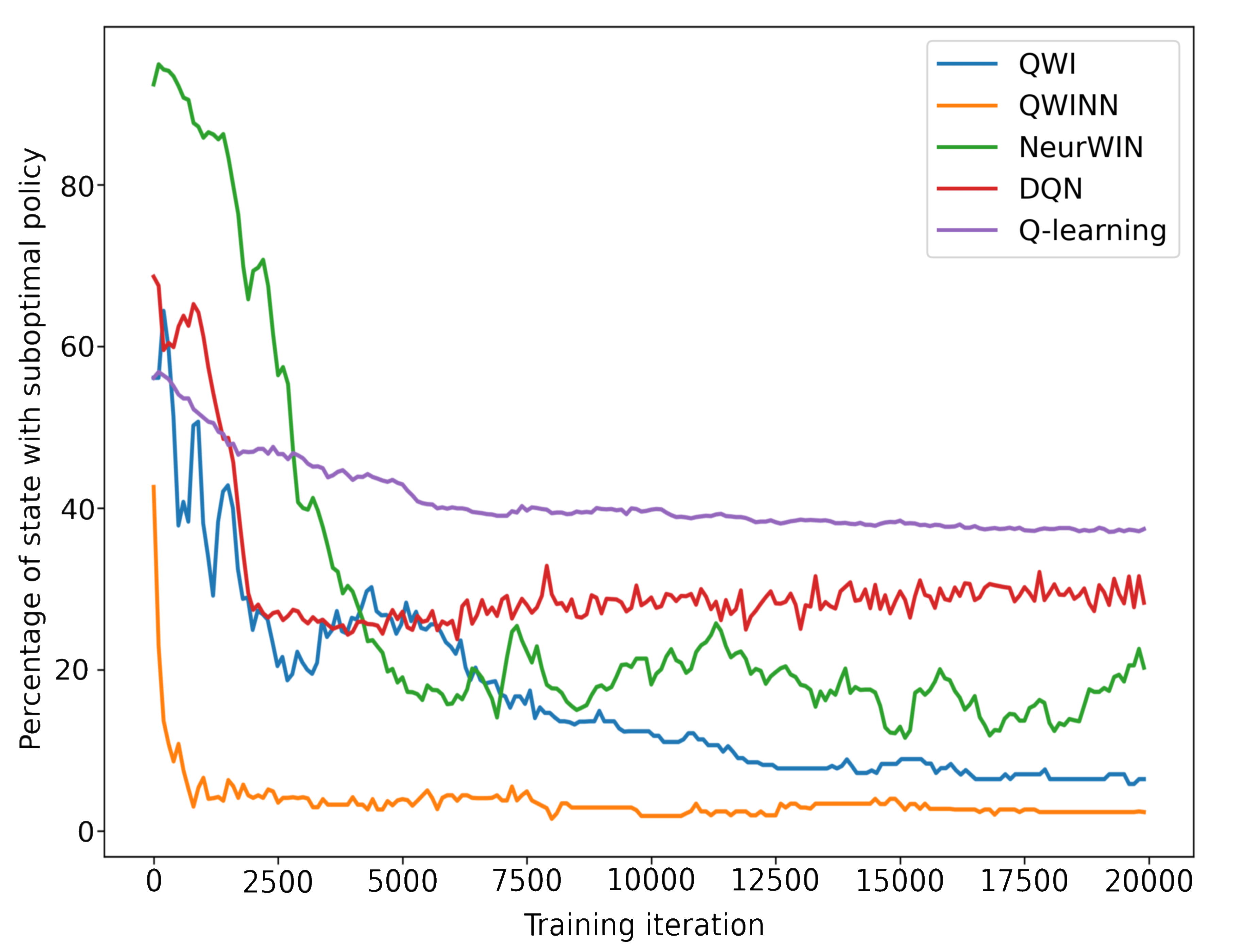}
        \label{fig_prob error restart - sub3}
    \end{subfigure}
    \hfill
    \begin{subfigure}{0.49\textwidth}
        \includegraphics[width=\textwidth]{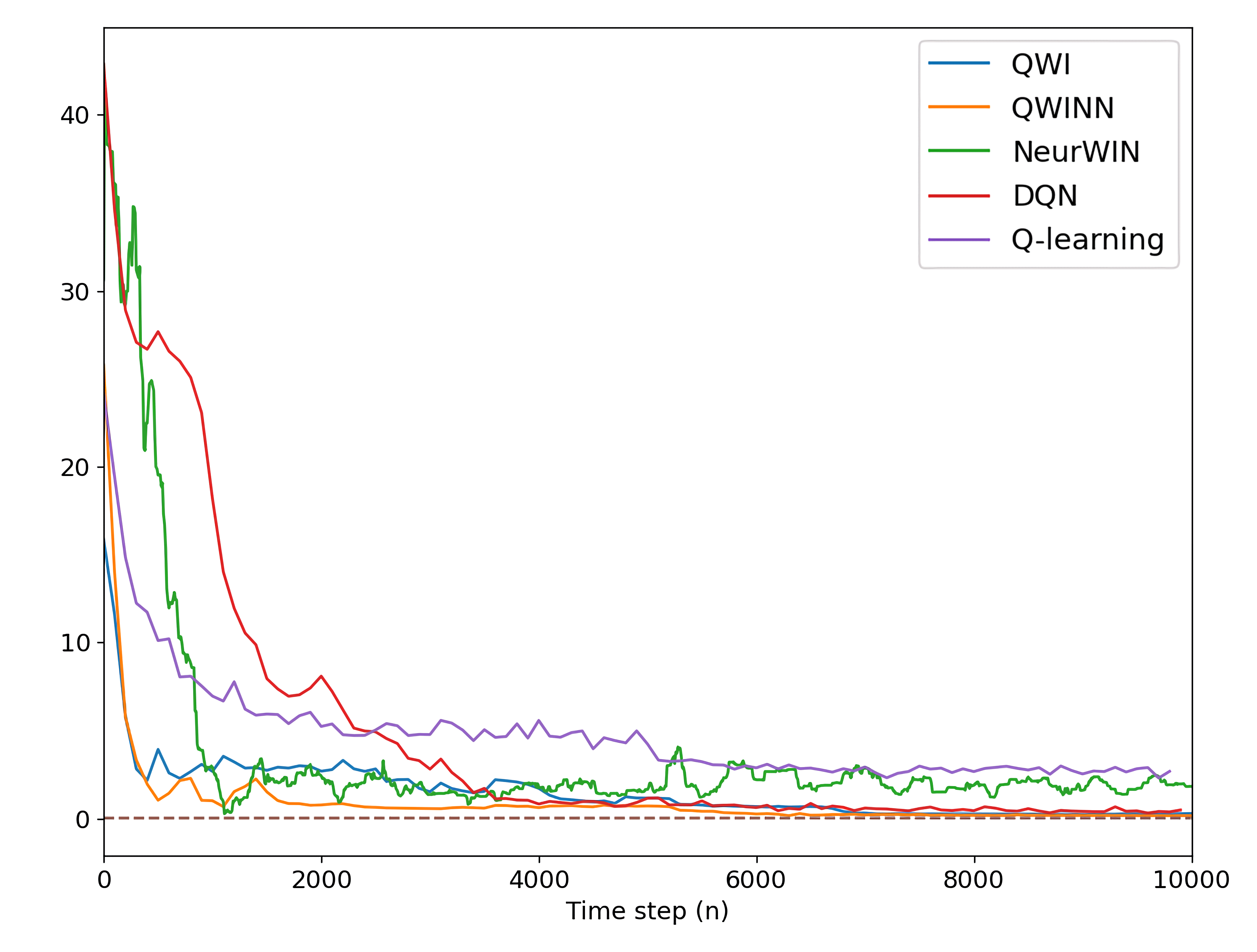}
        \label{fig_heterogeneous restart bellman - sub2}
    \end{subfigure}

    \caption{Performance graphs for the QWI, QWINN, NeurWIN, DQN and Q-learning algorithms: assignment of optimal policies in the heterogeneous ``restart'' problem \emph{(left)} and discounted rewards for the homogeneous ``deadline scheduling'' problem \emph{(right)}.}
    \label{fig:performance restart heterogeneous}
\end{figure}

In Figure~\ref{fig:performance restart heterogeneous} \emph{(right)} we plot the evolution of $BRE(P)$ for each of the algorithms during the training. We note that the state space is smaller than in the previous section, and as a consequence $Q$-learning converges faster, and that DQN essentially learns the optimal policy. Regarding QWI and QWINN, we observe that whereas the relative errors of QWI and QWINN are negligible and indistinguishable from DQN, NeurWIN incurs a tangible error of around $3\%$. The latter is due to lack of accuracy in the estimation of  Whittle index, which can induce a wrong ordering among the states. 
One way to show convergence to the correct index policy is to plot, for all possible combinations of states, what percentage of these do not perform the same action that Whittle index policy would render. In other words, during each instant of training, and for each possible combination of states among the different arms of the problem, we evaluate what action each algorithm's policy would perform against the Whittle index policy corresponding to that problem. In Figure \ref{fig:performance restart heterogeneous} \emph{(left)} we plot the percentage of state combinations in which the theoretical Whittle index action is not taken. 
For the restart problem with heterogeneous arms, using $N=3$, the total number of state combinations is $5^3$. 
As we can see, while QWINN reaches the theoretical Whittle index policy for most of the states in the first iterations, the error of QWI is around $10\%$ whereas that of NeurWIN is around $20\%$. The former is due to the difficulty of visiting the higher states of the problem while the latter is due to inaccurate index predictions, leading to sub-optimal policies especially in heterogeneous cases.

\subsection{Deadline scheduling problem}\label{sec_deadline scheduling}
In this section we consider the deadline scheduling problem studied in \cite{Nakhleh2021}, and for which the source code is publicly available. 
This problem, proposed in \cite{yu2018deadline}, has states formed by two different variables: The service time $B \in [0,9]$ and the deadline $T \in [0,12]$, leading to a total of $|S|=130$ states (although several of these states are not accessible by the Markov chain). Therefore, the variable $B$ represents the amount of workload pending to complete a certain job while the variable $T$ represents the remaining time available to perform it.
When there is no job at the $i$th position, the state is $(0,0)$, while otherwise it is $(T^i_n, B^i_n)$. In each iteration, the state transition $s^i_n = (T_n^i, B_n^i)$ depends on the action $a_n^i$ performed:
\[
    s^i_{n+1} = \begin{cases}
        (T_n^i-1, (B_n^i-a_n^i)^+) & \textrm{if } T_n^i > 1, \\
        (T,B) \textrm{ with prob. } Q(T,B) & \textrm{if } T_n^i \leq 1,
    \end{cases}
\]
where $b^+=\max(b,0)$.  When $T=1$, the deadline for completing the job ends, and the new state (including the empty state $(0,0)$) is chosen uniformly at random. The value of $B$, the workload to be completed, is only reduced if the action is active. 
If the scheduler reaches the state $(T=1, B>0)$, the job could not be finished on time, and an extra penalty $F(B_n^i - a_n^i)=0.2(B_n^i - a_n^i)^2$ is incurred. In addition, activating the arms involves a fixed cost $c=0.8 $, resulting in the following rewards:
\[
\begin{aligned}
    r_n^i(s_n^i, a_n^i, c) = 
    \begin{cases}
    (1 - c)a_n^i & \textrm{if } B_n^i > 0, T_n^i > 1 , \\
    (1 - c)a_n^i - F(B_n^i - a_n^i) & \textrm{if } B_n^i > 0, T_n^i = 1 ,\\
    0 & \textrm{otherwise}
    \end{cases} 
\end{aligned}
\]
In \cite{yu2018deadline} the authors provide an expression for Whittle index, given by:
\[
    \begin{aligned}
        \lambda(T,B,c) = 
        \begin{cases}
            0 & \textrm{if } B=0, \\
            1- c & \textrm{if } 1 \leq B \leq T-1, \\
            \begin{aligned}
                \gamma^{T-1}F(B-T+1) -
                \gamma^{T-1}F(B-T)  
                + 1- c 
            \end{aligned}
            & \textrm{if } T \leq B
        \end{cases}
    \end{aligned}
\]
In the following results we will consider the homogeneous case using $N=5$ and $M=2$, with a processing cost $c=0.8$ for all arms and a discount factor $\gamma=0.9$.

\begin{figure}[h]
    \centering
    \begin{subfigure}{0.49\textwidth}
        \includegraphics[width=\textwidth]{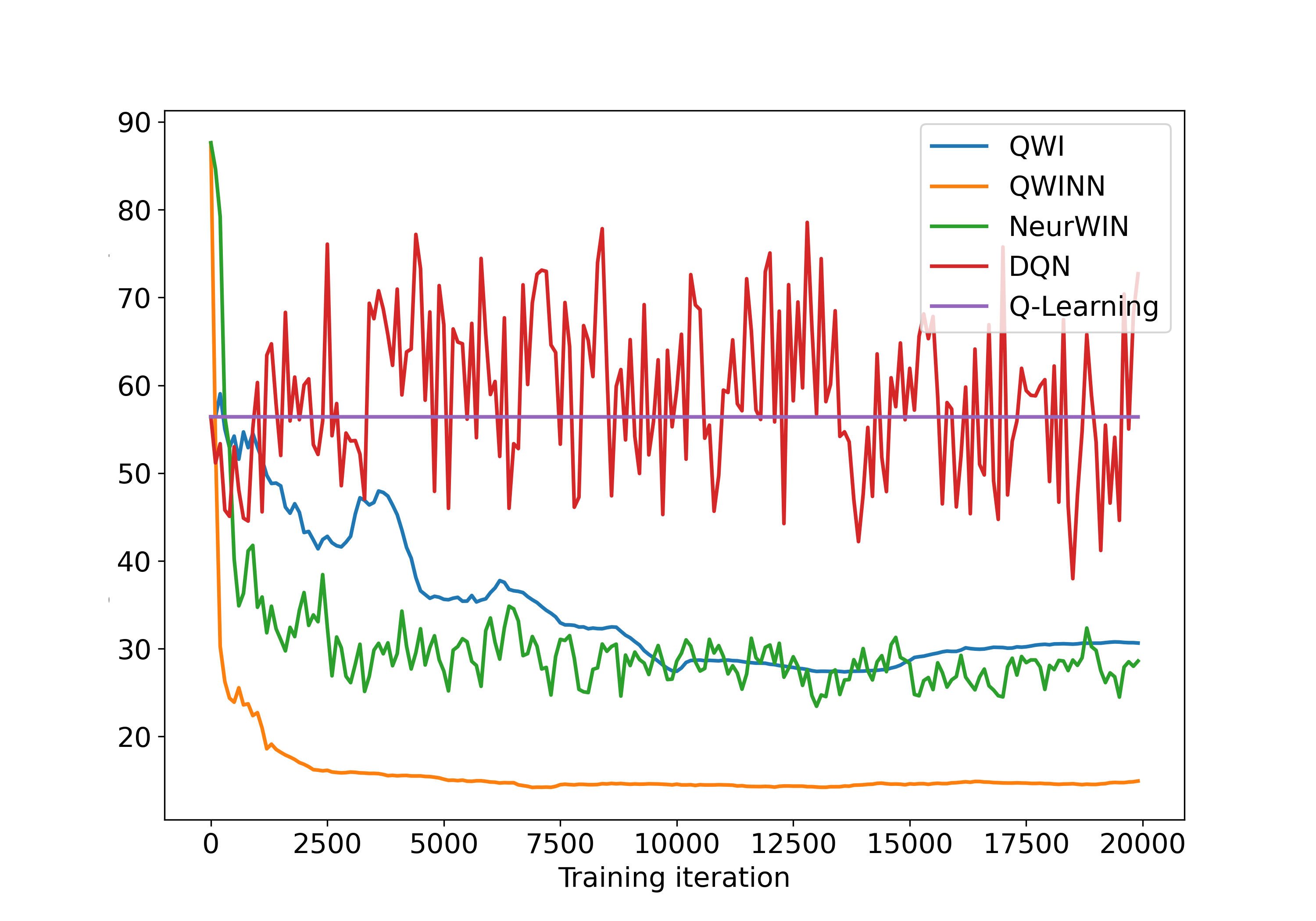}
        \caption{Percentage of states with suboptimal actions}
        \label{fig:deadline ordering - sub4}
    \end{subfigure}
    \hfill
    \begin{subfigure}{0.49\textwidth}
        \includegraphics[width=\textwidth]{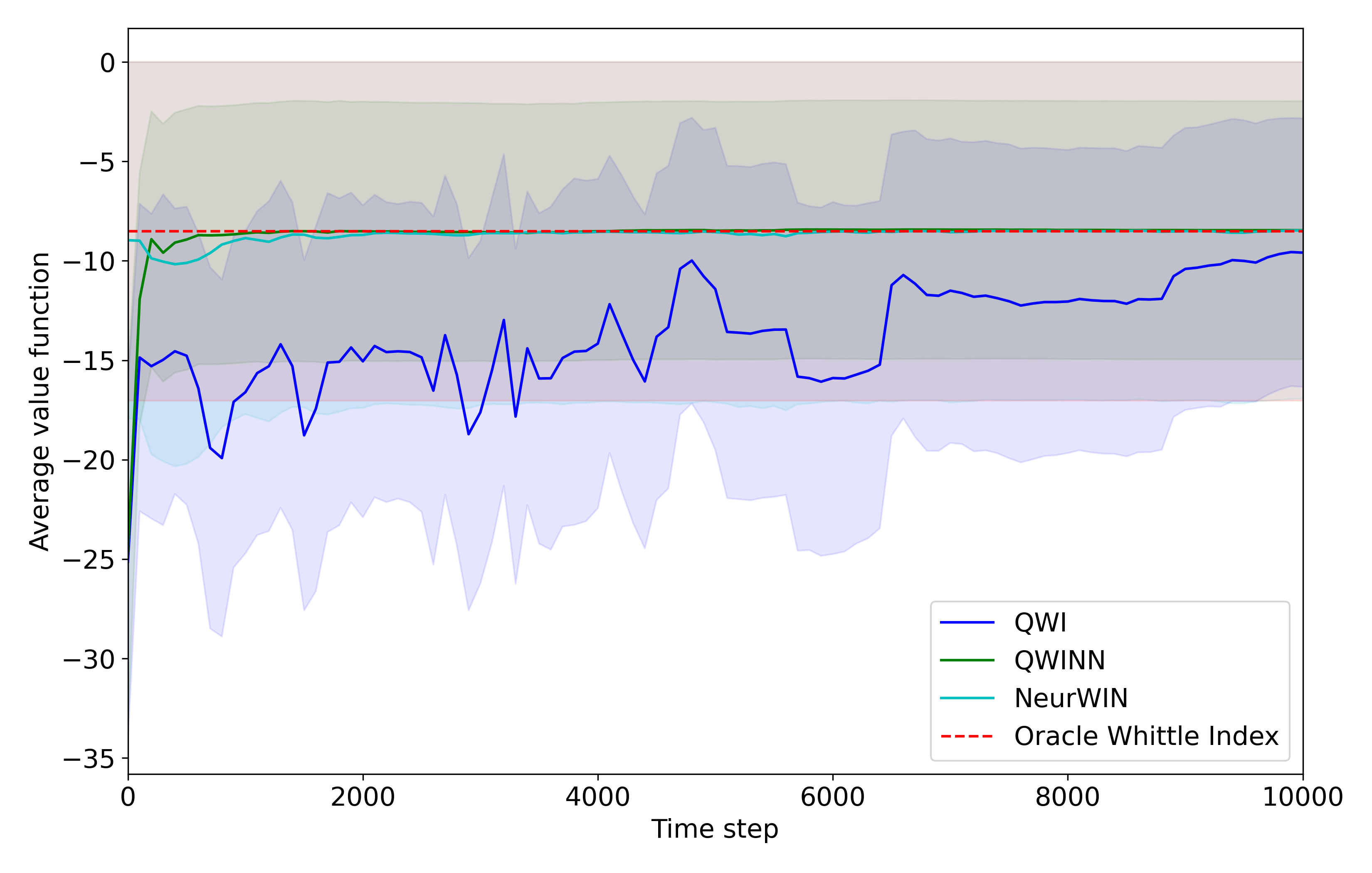}
        \caption{Average value function $N=5, M=2, |S|=130$}
        \label{fig:performance deadline homogeneous}
    \end{subfigure}
    \caption{Performance graphs for ``deadline scheduling problem'' using $N=5, M=2, |S|=130$.}
\end{figure}

In Figure \ref{fig:deadline ordering - sub4} we plot the percentage of state combinations in which the action of an algorithm would diverge from that prescribed by Whittle index policy. In other words, this happens when the algorithm fails to identify the best arm.
We observe that the QWINN is the algorithm with the smallest number of misorderings, and that both QWI and NeurWIN fail to identify the best arm in around 30\% of the states. 
By inspection, we have observed that QWI's estimates are poor for states with deadline $T>9$, which are rarely visited. This is a clear example of the limitations of a tabular algorithm when solving a problem with a large state space: by not having enough samples of these states, the tabular QWI algorithm is not able to obtain good predictions of the Whittle indices. On the other hand, we have not identified a particular set of states in which NeurWIN and QWINN fail. 

However, it is important to highlight that these misorderings do not necessarily imply a significant loss of performance. 
In Figure \ref{fig:performance deadline homogeneous}, we present the average value function for the QWI, QWINN, and NeurWIN algorithms during training, comparing them to the performance of the Whittle index Oracle policy for $N=5$ and $M=2$. As observed, while QWI takes longer to reach performance comparable to QWINN and NeurWIN, the latter two achieve close to identical performance to the Whittle index Oracle policy within the first few hundred iterations, indicating that the suboptimal states of NeurWIN do not significantly affect the evaluation of its policy. The performance of DQN and Q-learning has not been included in this analysis as neither is able to learn an efficient policy due to the size of the coupled state space.

\begin{figure}[h]
    \centering
    \begin{subfigure}{0.49\textwidth}
        \includegraphics[width=\textwidth]{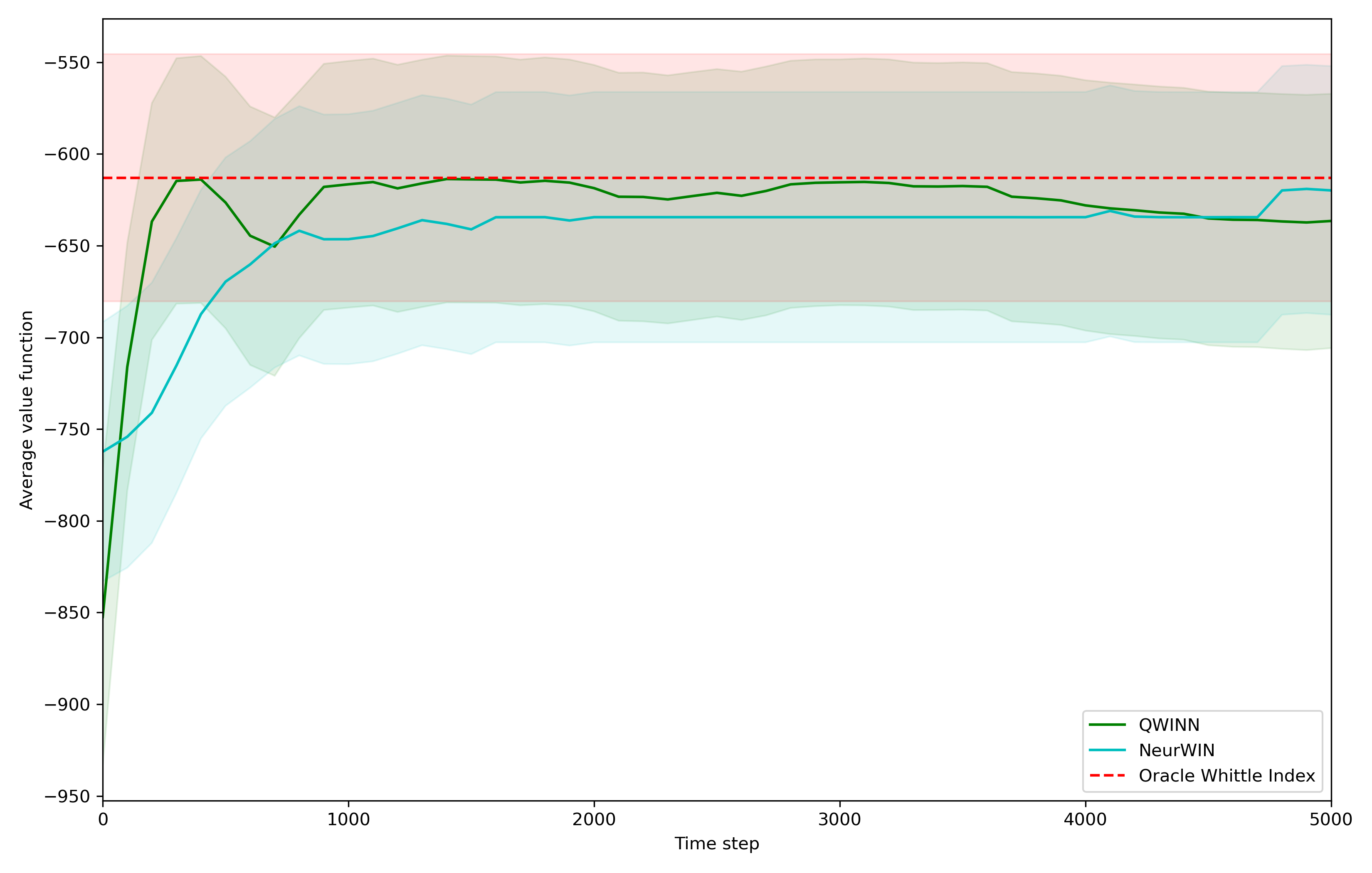}
        \caption{Reward evaluation, $N=100, M=20, |S|=255$}
        \label{fig:deadline large m20}
    \end{subfigure}
    \hfill
    \begin{subfigure}{0.49\textwidth}
        \includegraphics[width=\textwidth]{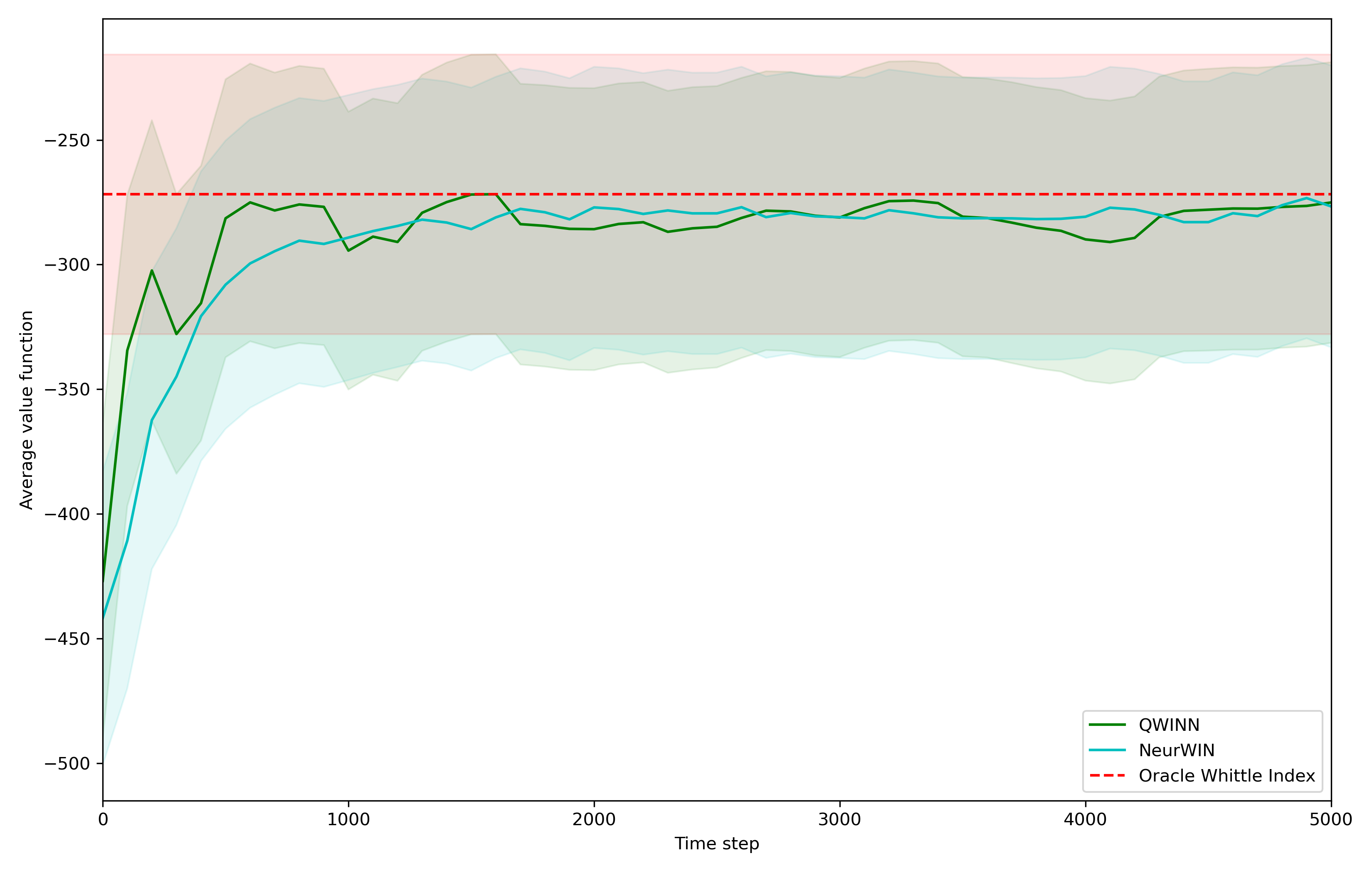}
        \caption{Reward evaluation, $N=100, M=50, |S|=255$}
        \label{fig:deadline large m50}
    \end{subfigure}
    \caption{Performance graphs for  ``deadline scheduling'' problem with $|S|=255$ states and $N=100$ arms using $M=\{20,50\}$ active arms}
    \label{fig:deadline large}
\end{figure}

In Figure \ref{fig:deadline large} we extend our analysis to a larger number of arms, $N=100$ while using $T\in [0, 17]$ and $B \in [0,15]$, resulting in a total of $|S|=255$ states, with different activation levels $M = \{20, 50\}$. Due to the increased state space, we limit our analysis to only QWINN and NeurWIN algorithms. In both Figure \ref{fig:deadline large m20} for $M=20$ and Figure \ref{fig:deadline large m50} for $M=50$, QWINN has a higher average performance with respect to NeurWIN, especially in the case of $M=20$, where QWINN's results are more stable. The difference between the two algorithms is smaller for the $M=50$ case as the agent is saturated by having to choose more arms, reducing the difference between the policies of the two algorithms.

\begin{figure}[h]
    \centering
    \includegraphics[width=0.49\textwidth]{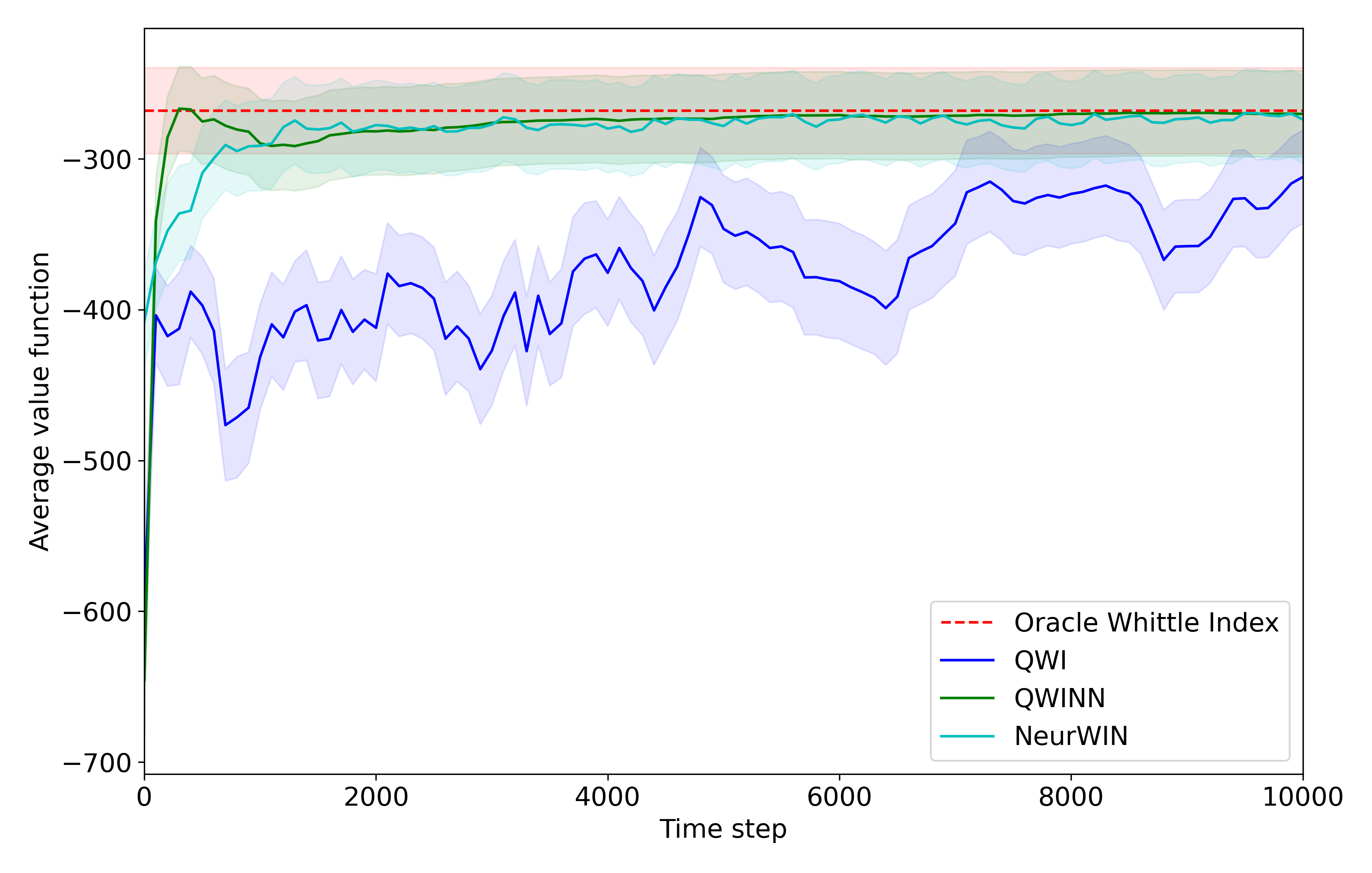}
    \caption{Reward evaluation of the NeurWIN and QWINN algorithms for a heterogeneous deadline scheduling problem with a state space of size $|S|=130$, $N=100$ arms, and $M=25$. Each of the four groups of 25 arms has a distinct cost (0.1, 0.3, 0.6, 0.8)}
    \label{fig:deadline heterogeneous value function}
\end{figure}

Lastly, let us examine the scenario of heterogeneous arms, where each arm has a state space of size $|S|=130$ and $N=100$. To differentiate the arms, we assign different activation cost values $c=\{0.1, 0.3, 0.6, 0.8\}$ for every 25 arms. Figure \ref{fig:deadline heterogeneous value function} illustrates the convergence of the algorithms. QWI falls short of reaching an optimal policy, while both QWINN and NeurWIN converge to Whittle's index Oracle policy. However, NeurWIN experiences performance perturbations caused by inaccurate index predictions, which become more pronounced with an increasing number of heterogeneous arms. In contrast, QWINN converges smoothly to the optimal policy without such performance fluctuations. It is worthy to note that while NeurWIN learns and then commits, QWINN operates in pure on-line mode.

\subsection{Circular problem}\label{sec:circular environment}
Finally, we consider here the ``circular'' problem, with a state space $S=\{0, 1, 2, 3\}$ \cite{fu2019towards}. In this problem,  with an active action the process remains in its current state with probability $0.6$, or increments positively with probability $0.4$. Similarly, with a passive action the process remains in its current state with probability $0.6$, or decrements negatively  with probability $0.4$. It is called circular, as an increment (decrement) from state 3 (0) brings the process to state 0 (3).
The reward function does not depend on the action performed, but only on the state, being $R(0)=-1, R(1)=R(2)=0, R(3)=1$. Using a discount parameter of $\gamma=0.9$, the theoretical values of Whittle index for each state are $\lambda(0)=-0.4390, \lambda(1)=0.4390, \lambda(2)=0.8652, \lambda(3)=-0.8652$, and we note that the best state is 1, and the second best 2.

\begin{figure}[h]
    \centering
    \begin{subfigure}{0.49\textwidth}
        \includegraphics[width=\textwidth]{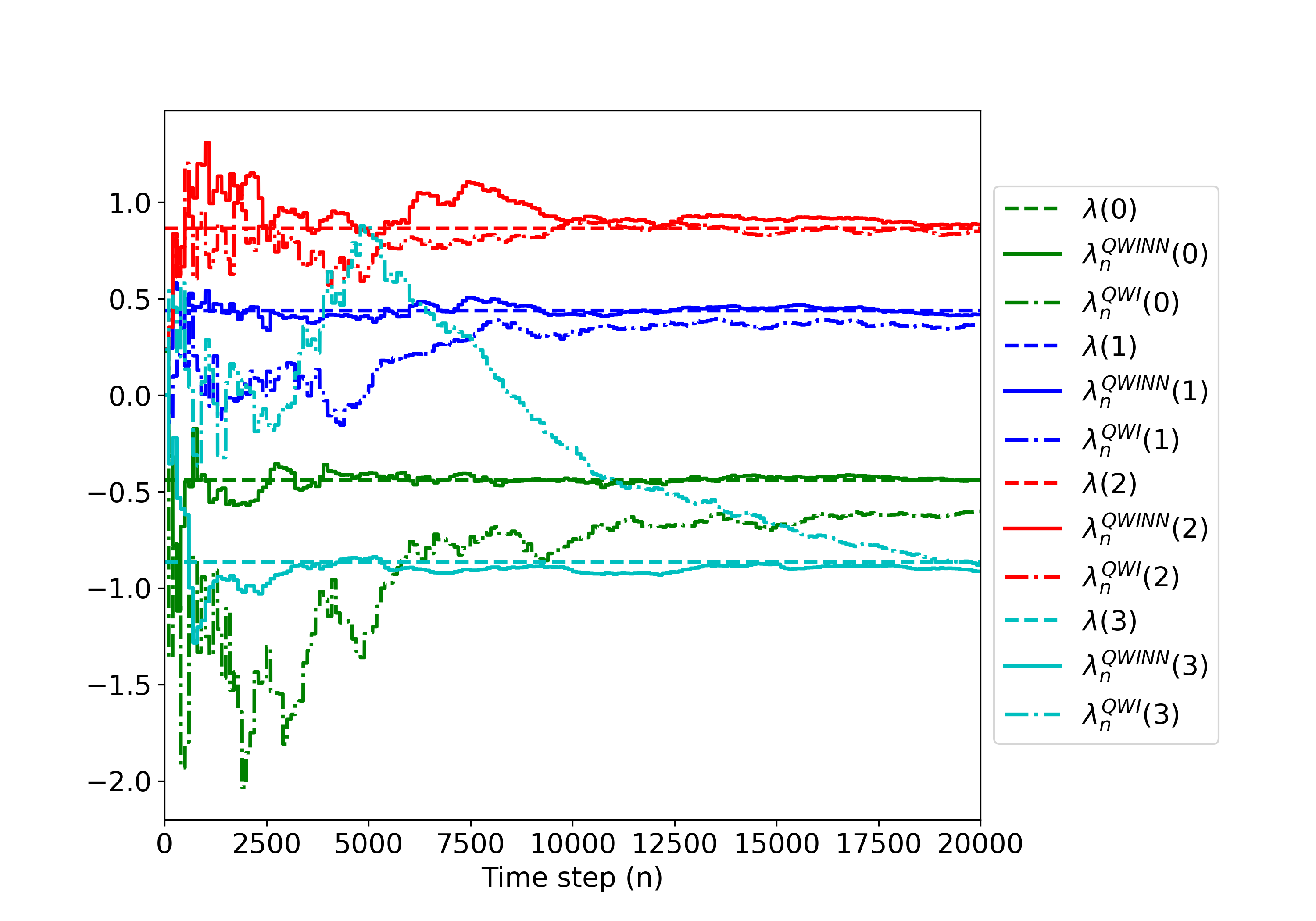}
        \caption{QWINN (solid lines) and QWI (dashdot lines)}
        \label{fig: circular QWI-QWINN index}
    \end{subfigure}
    \hfill
    \begin{subfigure}{0.49\textwidth}
        \includegraphics[width=\textwidth]{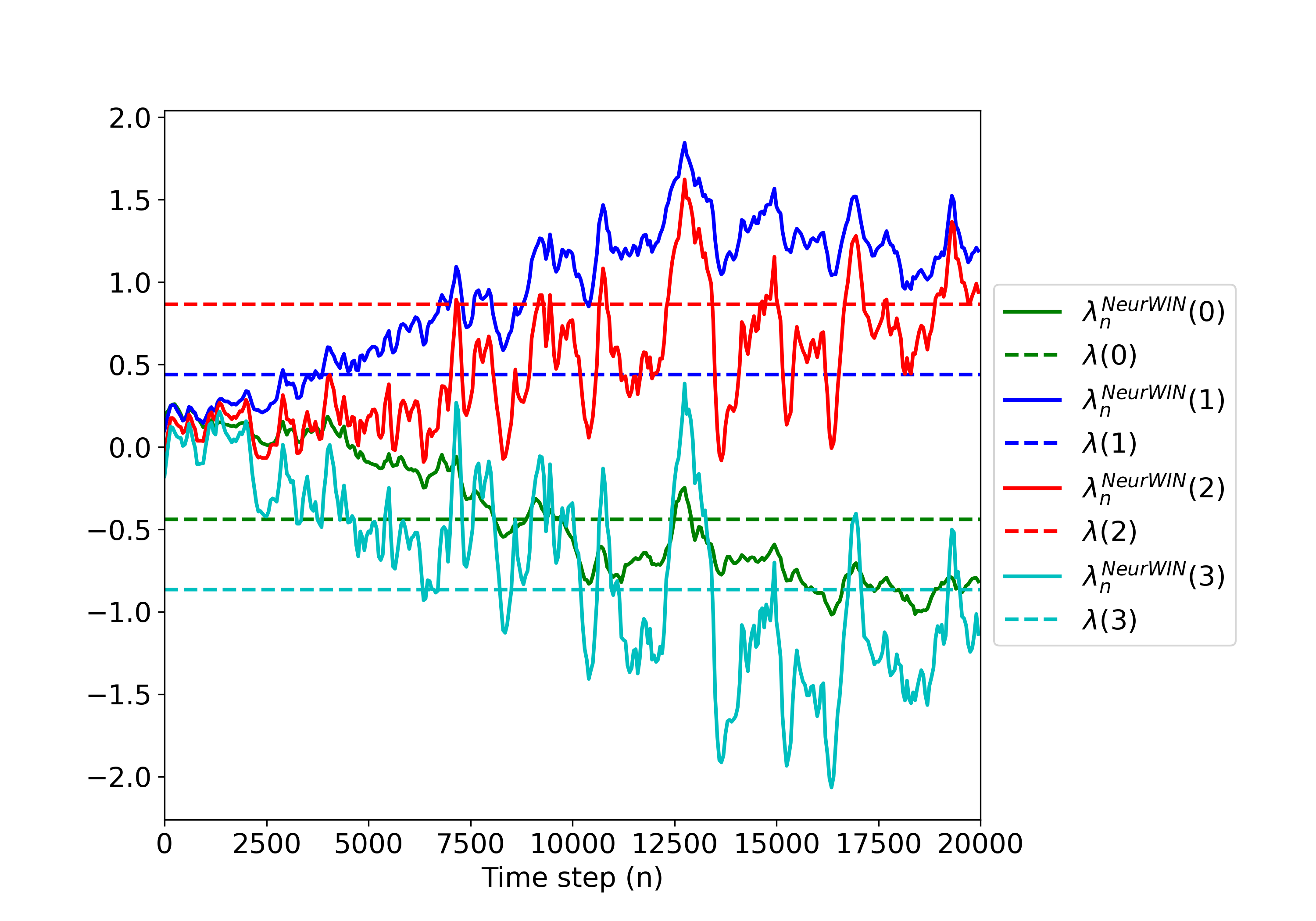}
        \caption{NeurWIN}
        \label{fig: circular neurwin index}
    \end{subfigure}
    \caption{Evolution of the Whittle index estimates for the circular problem}
    \label{fig:circular index}
\end{figure}

\begin{figure}[h]
    \centering
    \begin{subfigure}{0.49\textwidth}
        \includegraphics[width=\textwidth]{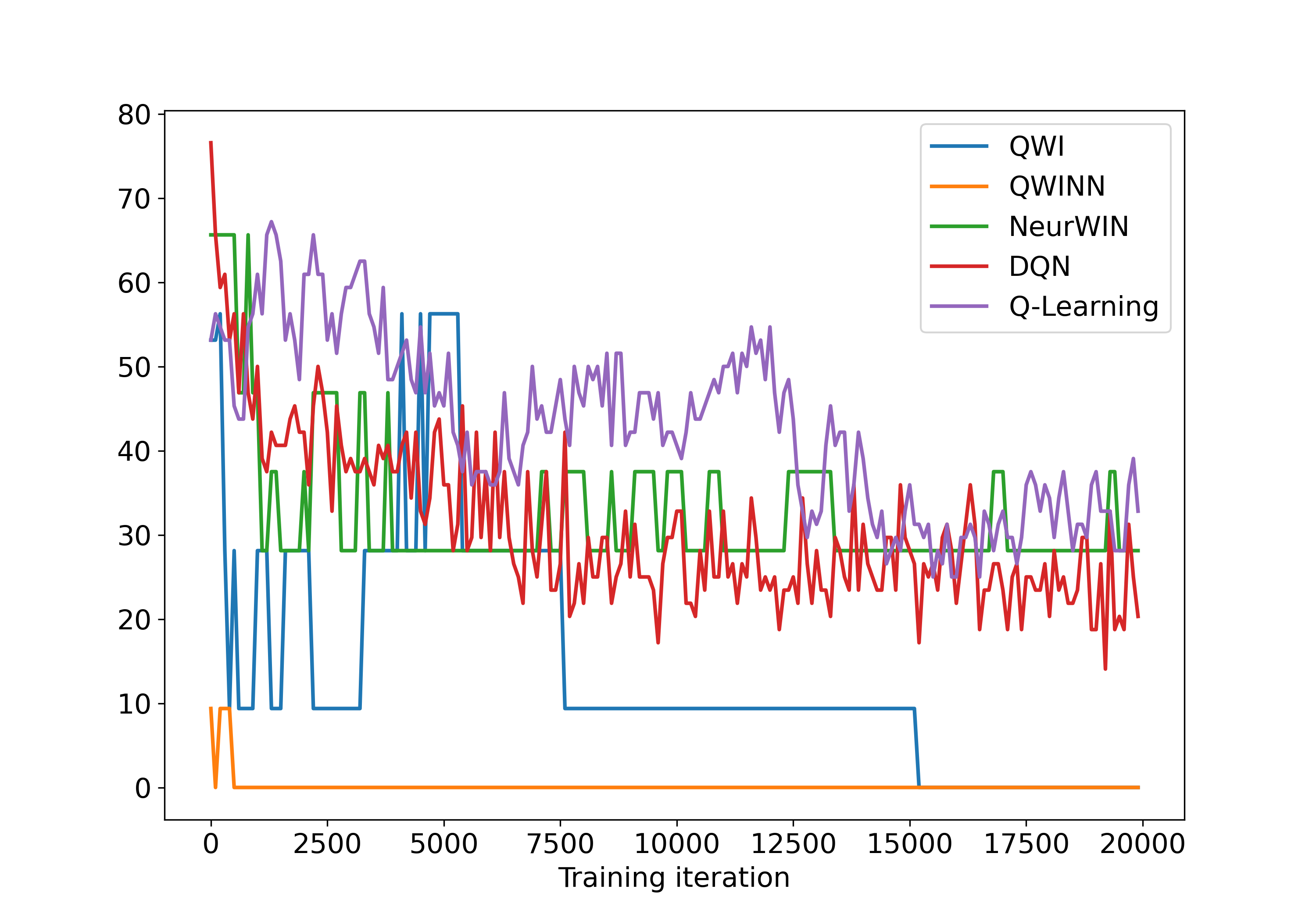}
        \caption{Percentage of states in which an optimal action is not performed in the ``circular'' problem with homogeneous arms}
        \label{fig:percentage error circular}
    \end{subfigure}
    \hfill
    \begin{subfigure}{0.49\textwidth}
        \includegraphics[width=\textwidth]{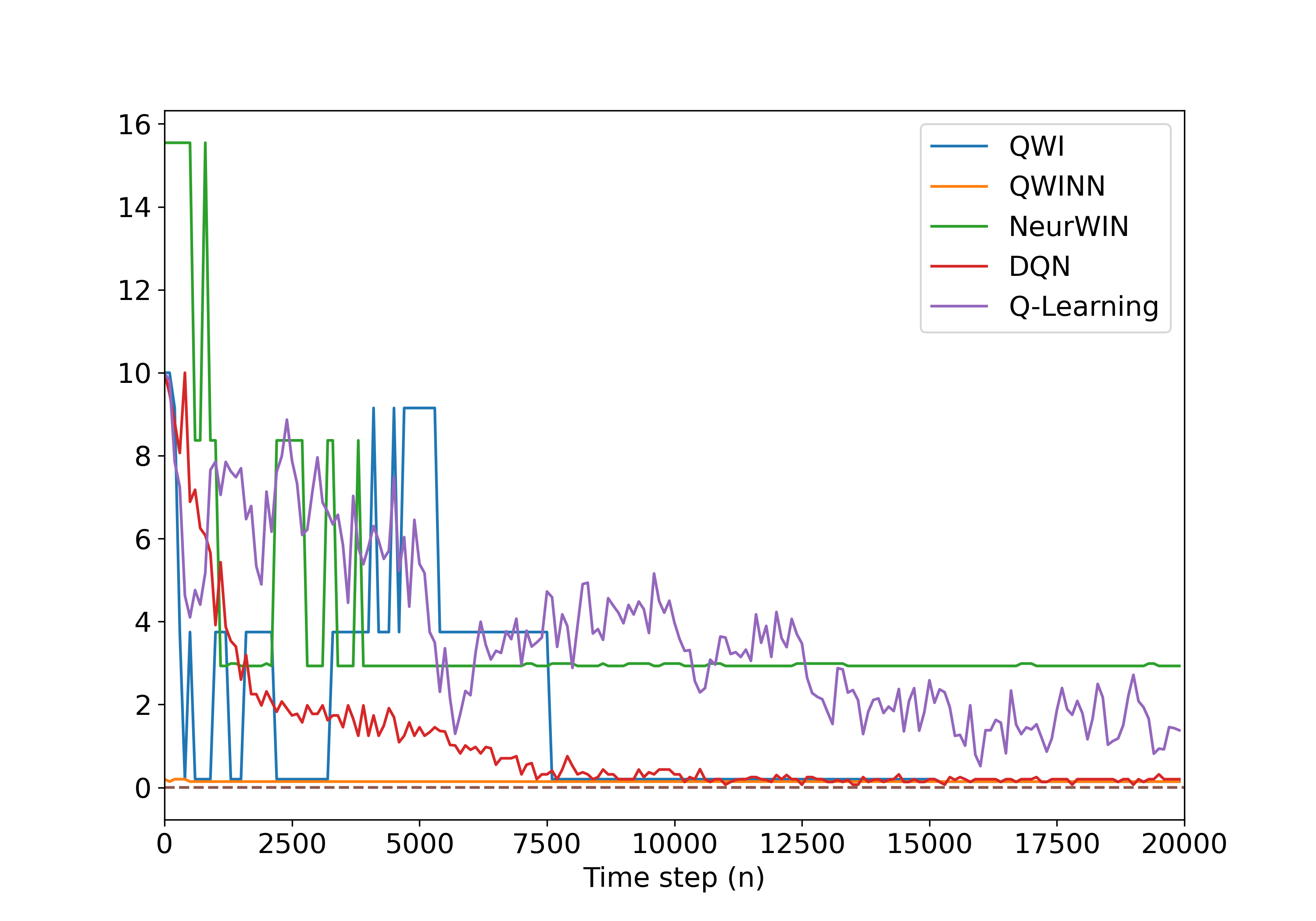}
        \caption{Bellman Relative Error $BRE(\pi_n^P)$,\\ $P\in \{\text{QWI, QWINN, NeurWIN, DQN, Q-Learning}\}$ }
        \label{fig:circular value function}
    \end{subfigure}
    \caption{Performance graphs for ``circular problem'' with $N=3, M=1, |S|=4$.}
    \label{fig:circular environment graphs}
\end{figure}

In Figure~\ref{fig:circular index}, we plot the Whittle index estimates for the QWI, QWINN and NeurWIN algorithms for this problem. We note that QWI and QWINN order correctly all states. but that with NeurWIN the estimate of the index of state 2 is consistently larger than that of state 1.  
We also note that QWINN's estimates converge the fastest. 
In Figure \ref{fig:percentage error circular} we plot the percentage of states in which the optimal action for the problem is not realized employing $N=3$ and $M=1$ with $\gamma=0.9$ in the QWI, QWINN, NeurWIN and classical DQN and Q-learning algorithms. Of all these algorithms, only QWINN and later QWI, are able to obtain an optimal action for all possible combinations of problem states. The rest of the algorithms remain with an error rate of 30\%, due in part to the difficulty of obtaining sufficient samples for all possible states in Q-learning, a poor regression in the case of DQN or a poor assignment of the indices in the case of NeurWIN, as can be seen in the Figure \ref{fig: circular neurwin index}.
The effect of these policies can be seen in Figure \ref{fig:circular value function}, where we plot the Bellman Relative Error $BRE(P)$ for QWI, QWINN, NeurWIN, DQN and Q-learning. As can be seen, none of the index heuristic algorithms achieves an optimal policy, since for this problem with this number of arms the optimal policy does not coincide with Whittle's index policy. Nevertheless, QWI and QWINN are able to achieve very good performance with respect to the optimal policy, especially QWINN which achieves this policy in the first few hundred iterations. On the other hand, due to NeurWIN's misordering of the indices, it shows a larger sub-optimality gap throughout the training. Q-learning and especially DQN perform very well, despite their poor state assignments, because the regret of these assignments is relatively small. 

\begin{figure}[h]
    \centering
    \begin{subfigure}[t]{0.5\textwidth}
        \centering
        \includegraphics[width=\linewidth]{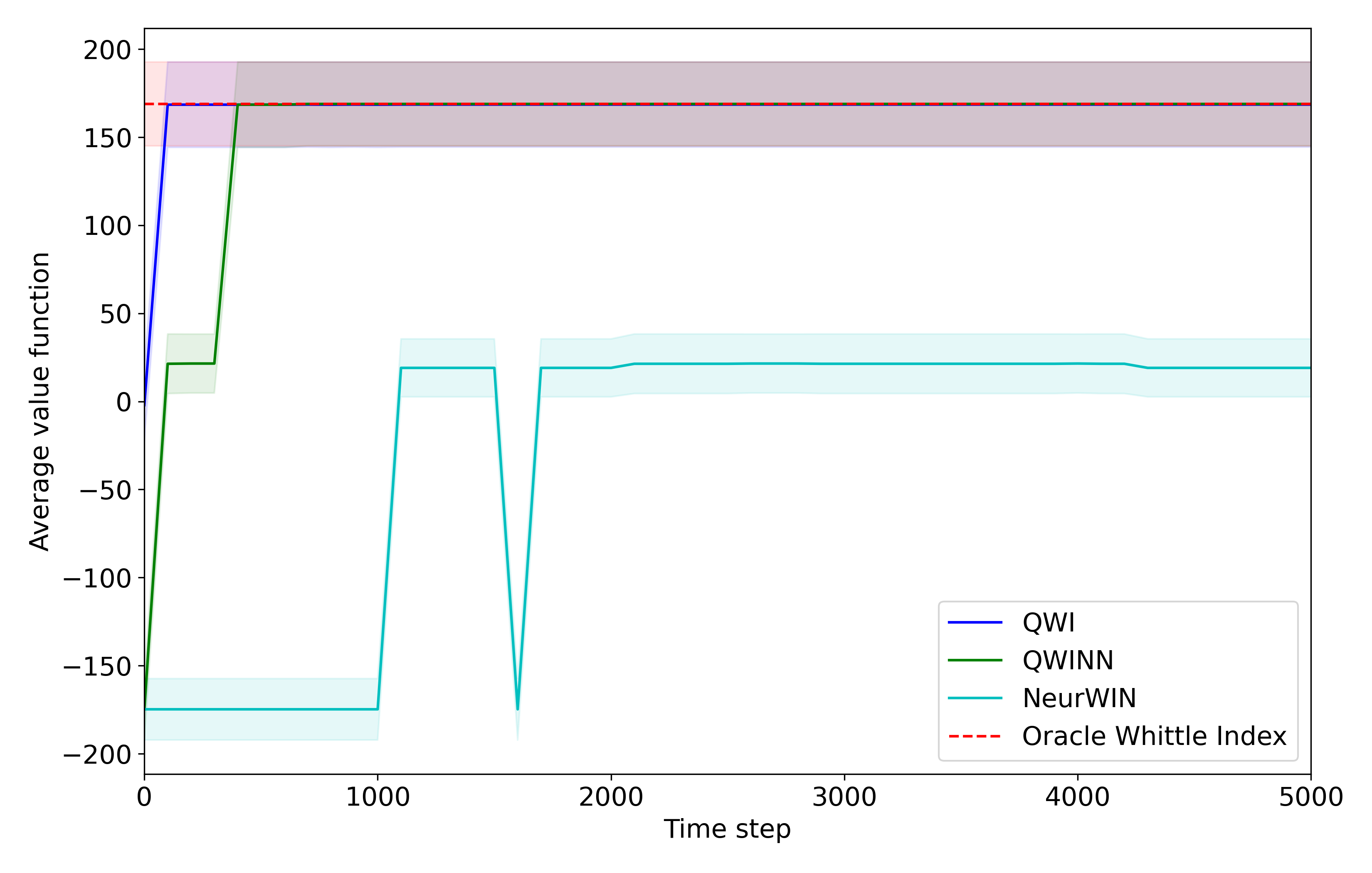}
        \caption{$N=100, M=20, |S|=4$}
        \label{fig:rewards circular - many arms}
    \end{subfigure}%
    \hfill
    \begin{subfigure}[t]{0.49\textwidth}
        \centering
        \includegraphics[width=\linewidth]{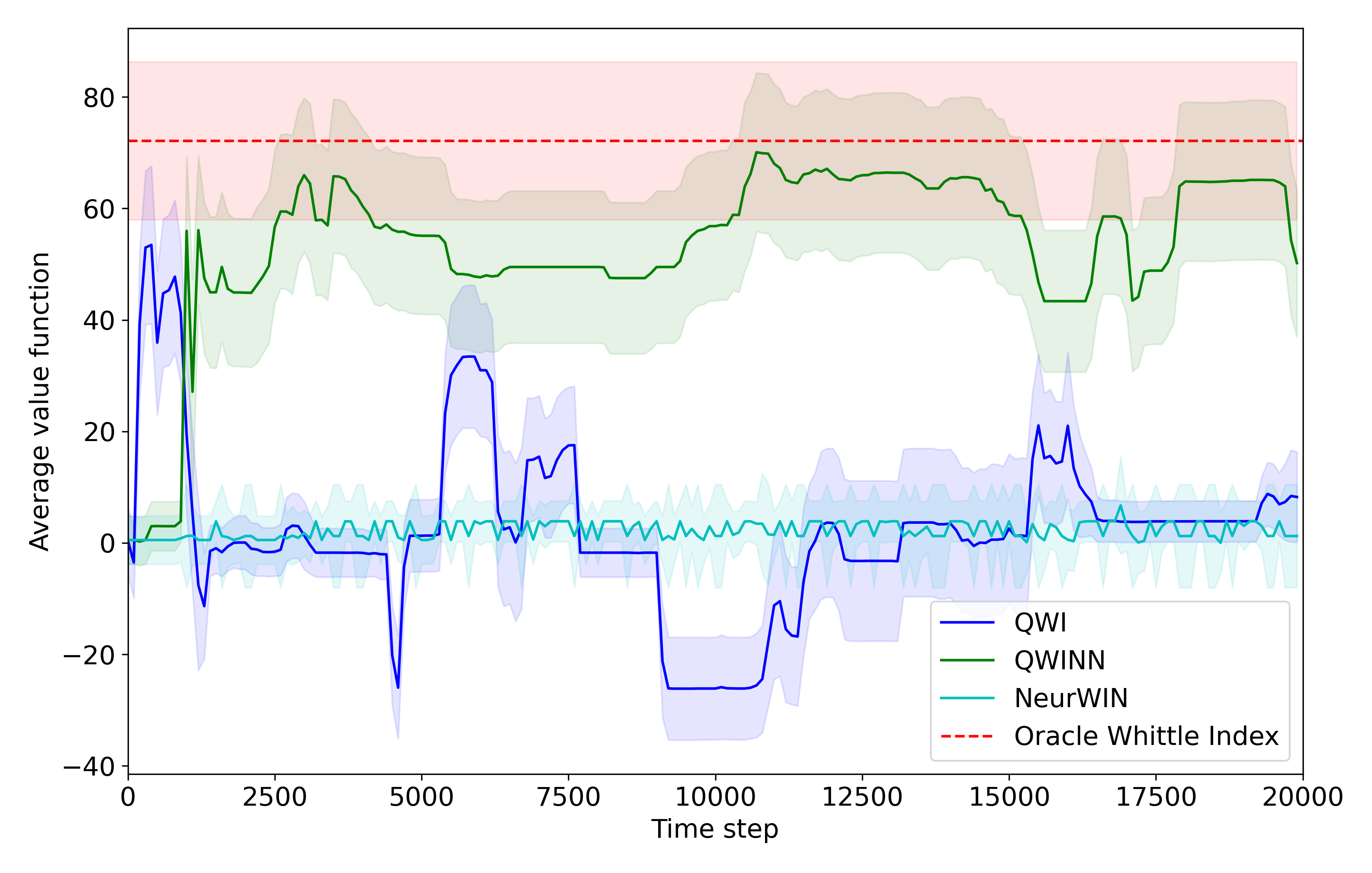}
        \caption{$N=100, M=50, |S|=50$}
        \label{fig:rewards circular - 50 active}
    \end{subfigure}%
    \hfill
    \begin{subfigure}[t]{0.49\textwidth}
        \centering
        \includegraphics[width=\linewidth]{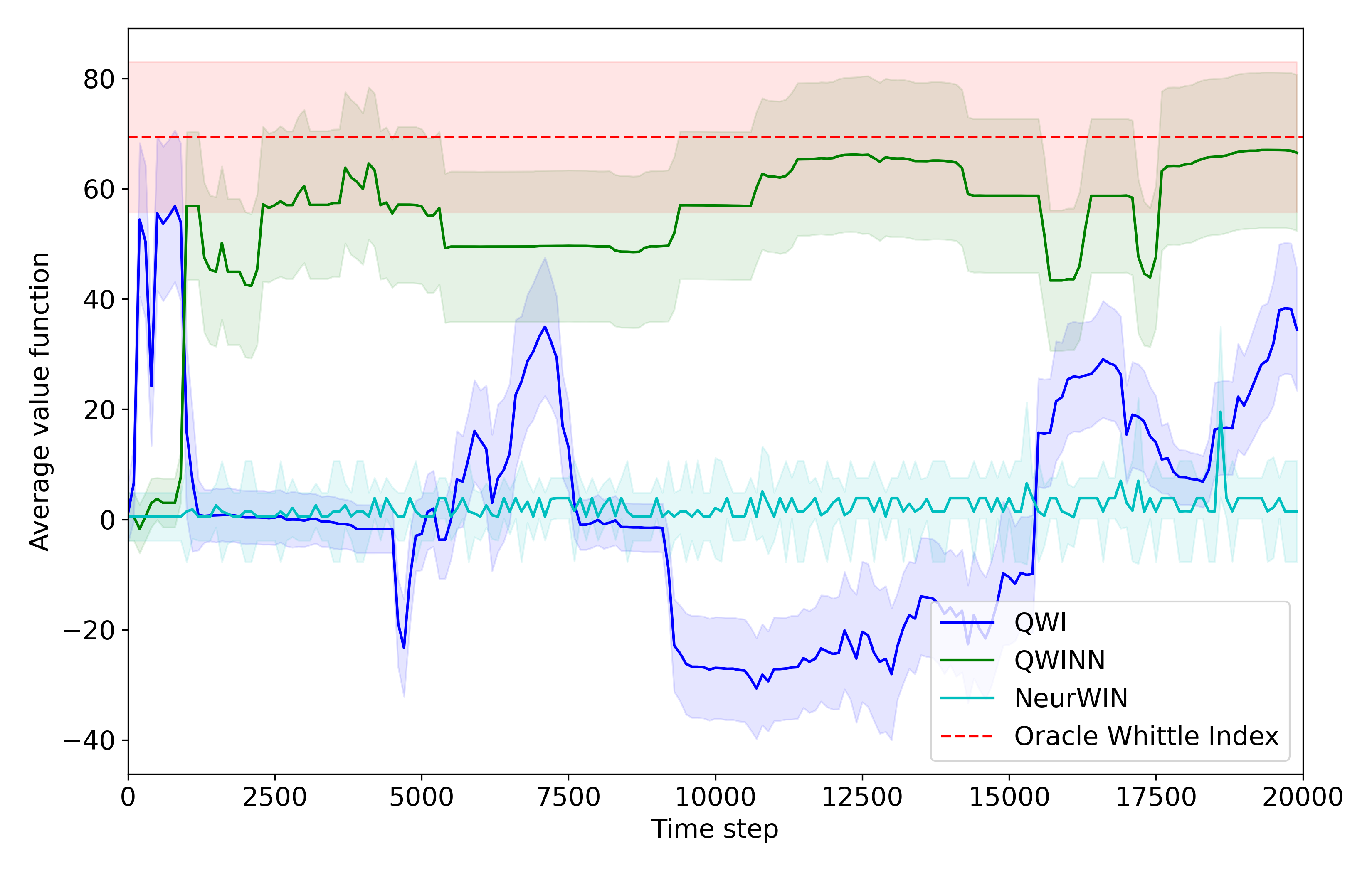}
        \caption{$N=100, M=70, |S|=50$}
        \label{fig: rewards circular - 70 active}
    \end{subfigure}
    \caption{Evaluation of the rewards obtained from the index policy for the circular dynamics problem for the QWI, QWINN and NeurWIN algorithms.}
    \label{fig:rewardscircular}
\end{figure}

Figure \ref{fig:rewardscircular} showcases the discounted value function achieved by the QWI, QWINN, and NeurWIN algorithms for different scenarios. In Figure \ref{fig:rewards circular - many arms}, we focus on the circular dynamics problem with a state space of |S|=4, N=100, and M=20. Figures \ref{fig:rewards circular - 50 active} and \ref{fig: rewards circular - 70 active} explore cases where we expand the state space to |S|=50, with only the first and last states having non-zero rewards (-1 and +1, respectively), along with N=100, M=50 (Figure \ref{fig:rewards circular - 50 active}) and N=100, M=70 (Figure \ref{fig: rewards circular - 70 active}).

In all these cases, NeurWIN struggles to generate an effective policy due to the incorrect ordering of its indices. Notably, in Figure \ref{fig:rewards circular - many arms}, where the state space is small, QWI achieves a Whittle index Oracle policy before QWINN. However, for the larger and sparser state spaces in Figures \ref{fig:rewards circular - 50 active} and \ref{fig: rewards circular - 70 active}, QWINN outperforms QWI by a significant margin.

\section{Conclusions}
In this paper we have developed two algorithms, QWI and QWINN, capable of learning the Whittle index for the total discounted criterion. We have presented analytical results about the asymptotic behavior of QWI as well as DQN and QWINN, the latter being the first of its kind. For QWINN, our analysis rigorously establishes the local convergence of the DQN algorithm under specific conditions. By strengthening assumptions on the Hessian's positive definiteness and applying stochastic approximation theory, we demonstrate that the DQN algorithm reliably converges to a neighborhood of a local minimum, particularly when iterates start close to this minimum and the algorithm incorporates a reset mechanism.
Through numerical simulations, we have compared the performance of QWI, QWINN with respect to other relevant algorithms, namely Q-larning, DQN and NeurWIN. Our results show that QWI, QWINN, NeurWIN are all more efficient than $Q$-learning in obtaining  policies with near-optimal performance.
QWI estimates very accurately the Whittle index in problems of small to moderate size. QWINN is able to accurately learn Whittle indices for problems with larger state spaces, and it is typically the fastest algorithm to converge. NeurWIN's estimates are not always accurate, but the latter does not necessarily imply a loss in performance. On the other hand, a simple problem like the circular problem shows that a wrong ordering can lead to a substantial performance degradation. 

An interesting and challenging direction for future research is the case of restless bandits with partial observations. The neural network based approach will be particularly useful there in order to mitigate the complexity of the belief state. Regret analysis in this framework is another research direction that needs to be addressed.

\section*{Acknowledgements}
We would like to express our gratitude to the reviewers and editors, who provided invaluable guidance to improve the paper.

F. Robledo and U. Ayesta have received funding from the Department of Education of the Basque Government through the Consolidated Research Group MATHMODE (IT1456-22).
Research partially supported by the French "Agence Nationale de la Recherche (ANR)" through the project ANR-22-CE25-0013-02 (ANR EPLER) and DST-Inria Cefipra project LION.
V. Borkar has received funding from the S. S. Bhatnagar Fellowship from Council of Scientific and Industrial Research, Government of India and Google Research.

\bibliographystyle{ACM-Reference-Format}
\bibliography{bib_new1.bib}


\begin{thebibliography}{28}


\ifx \showCODEN    \undefined \def \showCODEN     #1{\unskip}     \fi
\ifx \showDOI      \undefined \def \showDOI       #1{#1}\fi
\ifx \showISBNx    \undefined \def \showISBNx     #1{\unskip}     \fi
\ifx \showISBNxiii \undefined \def \showISBNxiii  #1{\unskip}     \fi
\ifx \showISSN     \undefined \def \showISSN      #1{\unskip}     \fi
\ifx \showLCCN     \undefined \def \showLCCN      #1{\unskip}     \fi
\ifx \shownote     \undefined \def \shownote      #1{#1}          \fi
\ifx \showarticletitle \undefined \def \showarticletitle #1{#1}   \fi
\ifx \showURL      \undefined \def \showURL       {\relax}        \fi
\providecommand\bibfield[2]{#2}
\providecommand\bibinfo[2]{#2}
\providecommand\natexlab[1]{#1}
\providecommand\showeprint[2][]{arXiv:#2}

\bibitem[Abounadi et~al\mbox{.}(2001)]%
        {abounadi2001learning}
\bibfield{author}{\bibinfo{person}{Jinane Abounadi}, \bibinfo{person}{Dimitrib Bertsekas}, {and} \bibinfo{person}{Vivek~S Borkar}.} \bibinfo{year}{2001}\natexlab{}.
\newblock \showarticletitle{Learning algorithms for Markov decision processes with average cost}.
\newblock \bibinfo{journal}{\emph{SIAM Journal on Control and Optimization}} \bibinfo{volume}{40}, \bibinfo{number}{3} (\bibinfo{year}{2001}), \bibinfo{pages}{681--698}.
\newblock
\urldef\tempurl%
\url{https://epubs.siam.org/doi/abs/10.1137/S0363012999361974?casa_token=1DzfRZBygwIAAAAA:1R6PB1DKopfx50oH5D9xAql38TAL8XbUL1TY3k8FJkc5NrkemLARZoiEd1paX3bA6zMAiHiklZWi-w}
\showURL{%
\tempurl}


\bibitem[Avrachenkov and Borkar(2022)]%
        {Avrachenkov2022}
\bibfield{author}{\bibinfo{person}{Konstantin~E. Avrachenkov} {and} \bibinfo{person}{Vivek~S. Borkar}.} \bibinfo{year}{2022}\natexlab{}.
\newblock \showarticletitle{Whittle index based {Q}-learning for restless bandits with average reward}.
\newblock \bibinfo{journal}{\emph{Automatica}}  \bibinfo{volume}{139} (\bibinfo{date}{May} \bibinfo{year}{2022}), \bibinfo{pages}{110186}.
\newblock
\showISSN{0005-1098}
\urldef\tempurl%
\url{https://doi.org/10.1016/j.automatica.2022.110186}
\showDOI{\tempurl}


\bibitem[Borkar et~al\mbox{.}(2008)]%
        {borkar2008stochastic}
\bibfield{author}{\bibinfo{person}{Vivek~S Borkar} {et~al\mbox{.}}} \bibinfo{year}{2008}\natexlab{}.
\newblock \showarticletitle{Stochastic approximation: a dynamical systems viewpoint}.
\newblock \bibinfo{journal}{\emph{Cambridge Books}} (\bibinfo{year}{2008}).
\newblock


\bibitem[Duff(1995)]%
        {duff1995q}
\bibfield{author}{\bibinfo{person}{Michael~O Duff}.} \bibinfo{year}{1995}\natexlab{}.
\newblock \showarticletitle{Q-learning for bandit problems}.
\newblock In \bibinfo{booktitle}{\emph{Machine Learning Proceedings 1995}}. \bibinfo{publisher}{Elsevier}, \bibinfo{pages}{209--217}.
\newblock
\urldef\tempurl%
\url{https://www.sciencedirect.com/science/article/pii/B9781558603776500347}
\showURL{%
\tempurl}


\bibitem[Fu et~al\mbox{.}(2019)]%
        {fu2019towards}
\bibfield{author}{\bibinfo{person}{Jing Fu}, \bibinfo{person}{Yoni Nazarathy}, \bibinfo{person}{Sarat Moka}, {and} \bibinfo{person}{Peter~G Taylor}.} \bibinfo{year}{2019}\natexlab{}.
\newblock \showarticletitle{Towards Q-learning the Whittle Index for Restless Bandits}. In \bibinfo{booktitle}{\emph{2019 Australian \& New Zealand Control Conference (ANZCC)}}. IEEE, \bibinfo{pages}{249--254}.
\newblock
\urldef\tempurl%
\url{https://doi.org/10.1109/anzcc47194.2019.8945748}
\showDOI{\tempurl}


\bibitem[Gibson et~al\mbox{.}(2021)]%
        {Gibson2021}
\bibfield{author}{\bibinfo{person}{Lachlan~J. Gibson}, \bibinfo{person}{Peter Jacko}, {and} \bibinfo{person}{Yoni Nazarathy}.} \bibinfo{year}{2021}\natexlab{}.
\newblock \showarticletitle{A {Novel} {Implementation} of {Q}-{Learning} for the {Whittle} {Index}}. In \bibinfo{booktitle}{\emph{Performance {Evaluation} {Methodologies} and {Tools}}} \emph{(\bibinfo{series}{Lecture {Notes} of the {Institute} for {Computer} {Sciences}, {Social} {Informatics} and {Telecommunications} {Engineering}})}, \bibfield{editor}{\bibinfo{person}{Qianchuan Zhao} {and} \bibinfo{person}{Li~Xia}} (Eds.). \bibinfo{publisher}{Springer International Publishing}, \bibinfo{address}{Cham}, \bibinfo{pages}{154--170}.
\newblock
\showISBNx{9783030925116}
\urldef\tempurl%
\url{https://doi.org/10.1007/978-3-030-92511-6_10}
\showDOI{\tempurl}


\bibitem[Gittins et~al\mbox{.}(2011)]%
        {GGW11}
\bibfield{author}{\bibinfo{person}{J.C. Gittins}, \bibinfo{person}{K. Glazebrook}, {and} \bibinfo{person}{R. Weber}.} \bibinfo{year}{2011}\natexlab{}.
\newblock \bibinfo{booktitle}{\emph{Multi-armed Bandit Allocation Indices}}.
\newblock \bibinfo{publisher}{Wiley}.
\newblock


\bibitem[Hsu(2018)]%
        {Hsu2018}
\bibfield{author}{\bibinfo{person}{Yu-Pin Hsu}.} \bibinfo{year}{2018}\natexlab{}.
\newblock \showarticletitle{Age of {Information}: {Whittle} {Index} for {Scheduling} {Stochastic} {Arrivals}}. In \bibinfo{booktitle}{\emph{2018 {IEEE} {International} {Symposium} on {Information} {Theory} ({ISIT})}}. \bibinfo{pages}{2634--2638}.
\newblock
\showISSN{2157-8117}
\urldef\tempurl%
\url{https://doi.org/10.1109/ISIT.2018.8437712}
\showDOI{\tempurl}
\newblock
\shownote{ISSN: 2157-8117}.


\bibitem[Killian et~al\mbox{.}(2021)]%
        {Killian2021}
\bibfield{author}{\bibinfo{person}{Jackson~A. Killian}, \bibinfo{person}{Arpita Biswas}, \bibinfo{person}{Sanket Shah}, {and} \bibinfo{person}{Milind Tambe}.} \bibinfo{year}{2021}\natexlab{}.
\newblock \showarticletitle{Q-{Learning} {Lagrange} {Policies} for {Multi}-{Action} {Restless} {Bandits}}. In \bibinfo{booktitle}{\emph{Proceedings of the 27th {ACM} {SIGKDD} {Conference} on {Knowledge} {Discovery} \& {Data} {Mining}}} \emph{(\bibinfo{series}{{KDD} '21})}. \bibinfo{publisher}{Association for Computing Machinery}, \bibinfo{address}{New York, NY, USA}, \bibinfo{pages}{871--881}.
\newblock
\showISBNx{9781450383325}
\urldef\tempurl%
\url{https://doi.org/10.1145/3447548.3467370}
\showDOI{\tempurl}


\bibitem[Lakshminarayanan and Bhatnagar(2017)]%
        {lakshminarayanan2017stability}
\bibfield{author}{\bibinfo{person}{Chandrashekar Lakshminarayanan} {and} \bibinfo{person}{Shalabh Bhatnagar}.} \bibinfo{year}{2017}\natexlab{}.
\newblock \showarticletitle{A stability criterion for two timescale stochastic approximation schemes}.
\newblock \bibinfo{journal}{\emph{Automatica}}  \bibinfo{volume}{79} (\bibinfo{year}{2017}), \bibinfo{pages}{108--114}.
\newblock
\urldef\tempurl%
\url{https://doi.org/10.1016/j.automatica.2016.12.014}
\showDOI{\tempurl}


\bibitem[Liu and Zhao(2010)]%
        {Liu2010}
\bibfield{author}{\bibinfo{person}{Keqin Liu} {and} \bibinfo{person}{Qing Zhao}.} \bibinfo{year}{2010}\natexlab{}.
\newblock \showarticletitle{Indexability of Restless Bandit Problems and Optimality of Whittle Index for Dynamic Multichannel Access}.
\newblock \bibinfo{journal}{\emph{IEEE Transactions on Information Theory}} \bibinfo{volume}{56}, \bibinfo{number}{11} (\bibinfo{date}{Nov.} \bibinfo{year}{2010}), \bibinfo{pages}{5547--5567}.
\newblock
\showISSN{1557-9654}
\urldef\tempurl%
\url{https://doi.org/10.1109/TIT.2010.2068950}
\showDOI{\tempurl}


\bibitem[Mnih et~al\mbox{.}(2015)]%
        {mnih2015human}
\bibfield{author}{\bibinfo{person}{Volodymyr Mnih}, \bibinfo{person}{Koray Kavukcuoglu}, \bibinfo{person}{David Silver}, \bibinfo{person}{Andrei~A Rusu}, \bibinfo{person}{Joel Veness}, \bibinfo{person}{Marc~G Bellemare}, \bibinfo{person}{Alex Graves}, \bibinfo{person}{Martin Riedmiller}, \bibinfo{person}{Andreas~K Fidjeland}, \bibinfo{person}{Georg Ostrovski}, {et~al\mbox{.}}} \bibinfo{year}{2015}\natexlab{}.
\newblock \showarticletitle{Human-level control through deep reinforcement learning}.
\newblock \bibinfo{journal}{\emph{nature}} \bibinfo{volume}{518}, \bibinfo{number}{7540} (\bibinfo{year}{2015}), \bibinfo{pages}{529--533}.
\newblock


\bibitem[Nakhleh et~al\mbox{.}(2021)]%
        {Nakhleh2021}
\bibfield{author}{\bibinfo{person}{Khaled Nakhleh}, \bibinfo{person}{Santosh Ganji}, \bibinfo{person}{Ping-Chun Hsieh}, \bibinfo{person}{I-Hong Hou}, {and} \bibinfo{person}{Srinivas Shakkottai}.} \bibinfo{year}{2021}\natexlab{}.
\newblock \showarticletitle{{NeurWIN}: {Neural} {Whittle} {Index} {Network} {For} {Restless} {Bandits} {Via} {Deep} {RL}}. In \bibinfo{booktitle}{\emph{Advances in {Neural} {Information} {Processing} {Systems}}}, Vol.~\bibinfo{volume}{34}. \bibinfo{publisher}{Curran Associates, Inc.}, \bibinfo{pages}{828--839}.
\newblock
\urldef\tempurl%
\url{https://proceedings.neurips.cc/paper/2021/hash/0768281a05da9f27df178b5c39a51263-Abstract.html}
\showURL{%
\tempurl}


\bibitem[Ni{\~n}o-Mora(2020)]%
        {NinoMora2020}
\bibfield{author}{\bibinfo{person}{Jos{\'e} Ni{\~n}o-Mora}.} \bibinfo{year}{2020}\natexlab{}.
\newblock \showarticletitle{A verification theorem for threshold-indexability of real-state discounted restless bandits}.
\newblock \bibinfo{journal}{\emph{Mathematics of Operations Research}} \bibinfo{volume}{45}, \bibinfo{number}{2} (\bibinfo{year}{2020}), \bibinfo{pages}{465--496}.
\newblock
\urldef\tempurl%
\url{https://doi.org/10.1287/moor.2019.0998}
\showDOI{\tempurl}


\bibitem[Ortner et~al\mbox{.}(2012)]%
        {ortner2012regret}
\bibfield{author}{\bibinfo{person}{Ronald Ortner}, \bibinfo{person}{Daniil Ryabko}, \bibinfo{person}{Peter Auer}, {and} \bibinfo{person}{R{\'e}mi Munos}.} \bibinfo{year}{2012}\natexlab{}.
\newblock \showarticletitle{Regret bounds for restless markov bandits}. In \bibinfo{booktitle}{\emph{International conference on algorithmic learning theory}}. Springer, \bibinfo{pages}{214--228}.
\newblock
\urldef\tempurl%
\url{https://link.springer.com/chapter/10.1007/978-3-642-34106-9_19}
\showURL{%
\tempurl}


\bibitem[Pagare et~al\mbox{.}(2023)]%
        {Pagare2023}
\bibfield{author}{\bibinfo{person}{Tejas Pagare}, \bibinfo{person}{Vivek Borkar}, {and} \bibinfo{person}{Konstantin Avrachenkov}.} \bibinfo{year}{2023}\natexlab{}.
\newblock \showarticletitle{Full Gradient Deep Reinforcement Learning for Average-Reward Criterion}. In \bibinfo{booktitle}{\emph{Learning for Dynamics and Control Conference}}. PMLR, \bibinfo{pages}{235--247}.
\newblock


\bibitem[Papadimitriou and Tsitsiklis(1999)]%
        {PapTsi1999}
\bibfield{author}{\bibinfo{person}{C.H. Papadimitriou} {and} \bibinfo{person}{J.N. Tsitsiklis}.} \bibinfo{year}{1999}\natexlab{}.
\newblock \showarticletitle{The Complexity of Optimal Queueing Network}.
\newblock \bibinfo{journal}{\emph{Mathematics of Operations Research}} \bibinfo{volume}{24}, \bibinfo{number}{2} (\bibinfo{year}{1999}), \bibinfo{pages}{293--305}.
\newblock


\bibitem[Puterman(2014)]%
        {Puterman2014}
\bibfield{author}{\bibinfo{person}{Martin~L. Puterman}.} \bibinfo{year}{2014}\natexlab{}.
\newblock \bibinfo{booktitle}{\emph{Markov {Decision} {Processes}: {Discrete} {Stochastic} {Dynamic} {Programming}}}.
\newblock \bibinfo{publisher}{John Wiley \& Sons}.
\newblock
\showISBNx{9781118625873}
\newblock
\shownote{Google-Books-ID: VvBjBAAAQBAJ}.


\bibitem[Sutton and Barto(2018)]%
        {sutton2018reinforcement}
\bibfield{author}{\bibinfo{person}{Richard~S Sutton} {and} \bibinfo{person}{Andrew~G Barto}.} \bibinfo{year}{2018}\natexlab{}.
\newblock \bibinfo{booktitle}{\emph{Reinforcement learning: An introduction}}.
\newblock \bibinfo{publisher}{MIT press}.
\newblock


\bibitem[Van~Hasselt(2010)]%
        {Hasselt2010}
\bibfield{author}{\bibinfo{person}{Hado Van~Hasselt}.} \bibinfo{year}{2010}\natexlab{}.
\newblock \showarticletitle{Double {Q}-learning}. In \bibinfo{booktitle}{\emph{Advances in {Neural} {Information} {Processing} {Systems}}}, Vol.~\bibinfo{volume}{23}. \bibinfo{publisher}{Curran Associates, Inc.}
\newblock
\urldef\tempurl%
\url{https://proceedings.neurips.cc/paper/2010/hash/091d584fced301b442654dd8c23b3fc9-Abstract.html}
\showURL{%
\tempurl}


\bibitem[Van~Hasselt et~al\mbox{.}(2016)]%
        {van2016deep}
\bibfield{author}{\bibinfo{person}{Hado Van~Hasselt}, \bibinfo{person}{Arthur Guez}, {and} \bibinfo{person}{David Silver}.} \bibinfo{year}{2016}\natexlab{}.
\newblock \showarticletitle{Deep Reinforcement Learning with Double Q-Learning}. In \bibinfo{booktitle}{\emph{Proceedings of the AAAI conference on artificial intelligence}}, Vol.~\bibinfo{volume}{30}.
\newblock
\urldef\tempurl%
\url{https://ojs.aaai.org/index.php/AAAI/article/view/10295}
\showURL{%
\tempurl}


\bibitem[Verloop(2016)]%
        {Verloop2016}
\bibfield{author}{\bibinfo{person}{I.~M. Verloop}.} \bibinfo{year}{2016}\natexlab{}.
\newblock \showarticletitle{Asymptotically optimal priority policies for indexable and nonindexable restless bandits}.
\newblock \bibinfo{journal}{\emph{The Annals of Applied Probability}} \bibinfo{volume}{26}, \bibinfo{number}{4} (\bibinfo{date}{Aug.} \bibinfo{year}{2016}), \bibinfo{pages}{1947--1995}.
\newblock
\showISSN{1050-5164, 2168-8737}
\urldef\tempurl%
\url{https://doi.org/10.1214/15-AAP1137}
\showDOI{\tempurl}


\bibitem[Wang et~al\mbox{.}(2020)]%
        {Wang2020}
\bibfield{author}{\bibinfo{person}{Siwei Wang}, \bibinfo{person}{Longbo Huang}, {and} \bibinfo{person}{John C.~S. Lui}.} \bibinfo{year}{2020}\natexlab{}.
\newblock \showarticletitle{Restless-{UCB}, an {Efficient} and {Low}-complexity {Algorithm} for {Online} {Restless} {Bandits}}. In \bibinfo{booktitle}{\emph{Advances in {Neural} {Information} {Processing} {Systems}}}, Vol.~\bibinfo{volume}{33}. \bibinfo{publisher}{Curran Associates, Inc.}, \bibinfo{pages}{11878--11889}.
\newblock
\urldef\tempurl%
\url{https://proceedings.neurips.cc/paper/2020/hash/89ae0fe22c47d374bc9350ef99e01685-Abstract.html}
\showURL{%
\tempurl}


\bibitem[Watkins and Dayan(1992)]%
        {watkins1992q}
\bibfield{author}{\bibinfo{person}{Christopher~JCH Watkins} {and} \bibinfo{person}{Peter Dayan}.} \bibinfo{year}{1992}\natexlab{}.
\newblock \showarticletitle{Q-learning}.
\newblock \bibinfo{journal}{\emph{Machine learning}} \bibinfo{volume}{8}, \bibinfo{number}{3-4} (\bibinfo{year}{1992}), \bibinfo{pages}{279--292}.
\newblock
\urldef\tempurl%
\url{https://link.springer.com/content/pdf/10.1007/BF00992698.pdf}
\showURL{%
\tempurl}


\bibitem[Weber and Weiss(1990)]%
        {weber1990index}
\bibfield{author}{\bibinfo{person}{Richard~R Weber} {and} \bibinfo{person}{Gideon Weiss}.} \bibinfo{year}{1990}\natexlab{}.
\newblock \showarticletitle{On an index policy for restless bandits}.
\newblock \bibinfo{journal}{\emph{Journal of applied probability}} (\bibinfo{year}{1990}), \bibinfo{pages}{637--648}.
\newblock
\urldef\tempurl%
\url{https://doi.org/10.2307/3214547}
\showDOI{\tempurl}


\bibitem[Whittle(1988)]%
        {whittle1988restless}
\bibfield{author}{\bibinfo{person}{Peter Whittle}.} \bibinfo{year}{1988}\natexlab{}.
\newblock \showarticletitle{Restless bandits: Activity allocation in a changing world}.
\newblock \bibinfo{journal}{\emph{Journal of Applied Probability}} \bibinfo{volume}{25}, \bibinfo{number}{A} (\bibinfo{date}{Jan.} \bibinfo{year}{1988}), \bibinfo{pages}{287--298}.
\newblock
\showISSN{0021-9002, 1475-6072}
\urldef\tempurl%
\url{https://doi.org/10.2307/3214163}
\showDOI{\tempurl}


\bibitem[Xiong and Li(2023)]%
        {xiong2023finite}
\bibfield{author}{\bibinfo{person}{Guojun Xiong} {and} \bibinfo{person}{Jian Li}.} \bibinfo{year}{2023}\natexlab{}.
\newblock \showarticletitle{Finite-Time Analysis of Whittle Index based Q-Learning for Restless Multi-Armed Bandits with Neural Network Function Approximation}. In \bibinfo{booktitle}{\emph{Thirty-seventh Conference on Neural Information Processing Systems}}.
\newblock


\bibitem[Yu et~al\mbox{.}(2018)]%
        {yu2018deadline}
\bibfield{author}{\bibinfo{person}{Zhe Yu}, \bibinfo{person}{Yunjian Xu}, {and} \bibinfo{person}{Lang Tong}.} \bibinfo{year}{2018}\natexlab{}.
\newblock \showarticletitle{Deadline scheduling as restless bandits}.
\newblock \bibinfo{journal}{\emph{IEEE Trans. Automat. Control}} \bibinfo{volume}{63}, \bibinfo{number}{8} (\bibinfo{year}{2018}), \bibinfo{pages}{2343--2358}.
\newblock
\urldef\tempurl%
\url{https://ieeexplore.ieee.org/abstract/document/8295041/}
\showURL{%
\tempurl}


\end{thebibliography}


\newpage

\appendix
\section{Proof of Theorem~\ref{thmQWI}}\label{sec:proof of convergence annex}

The following assumptions are required:
\begin{enumerate}
    \item[(C1)] Step sizes $\{\alpha(n)\}$ satisfy, for $x \in (0,1)$,
    \[
    \begin{aligned}
        \sup_n & \frac{\alpha(\lfloor xn \rfloor)}{\alpha(n)} < \infty,
        \\
        \sup_{y \in [x,1]} & \left| \frac{\sum_{m=0}^{\lfloor yn \rfloor} \alpha(m)}{\sum_{m=0}^{n} \alpha(m)} - 1 \right| \rightarrow 0
    \end{aligned}
    \]
    \item[(C2)] The problem is Whittle indexable, that is, given the subset of states $\Pi(\lambda)$ where the optimal action is the passive action due to the extra reward $\lambda$ of this action, the number of states within this subset $\Pi(\lambda)$ must grow monotonically as $\lambda$ increases.
\end{enumerate}

 We will follow the general scheme proposed in \cite{lakshminarayanan2017stability} and \cite{borkar2008stochastic}. Consider a generalised version of equations (\ref{eq_q-learning with x}) and (\ref{eq:learingW}) such that:
\begin{align}
    x_{n+1}= x_n + a(n) \left[h(x_n, y_n) + M_{n+1}^{(1)} \right] \label{eq_update x}\\
    y_{n+1}= y_n + b(n) \left[g(x_n, y_n) + M_{n+1}^{(2)} \right] \label{eq_update y}
\end{align}
where (\ref{eq_update x}) represents Equation (\ref{eq_q-learning with x}), the fast time-scale where we compute the $Q$-values, and (\ref{eq_update y}) represents Equation (\ref{eq:learingW}), the slower time-scale in which the Whittle indices are updated. In these equations, the functions $h$ and $g$ are continuous Lipschitz functions, the $M_n$ are martingale difference sequences representing noise terms, and $a(n)$ and $b(n)$ are step-size terms satisfying $\frac{b(n)}{a(n)} \rightarrow 0$ as $n \rightarrow \infty$, in addition to the usual conditions $\sum_na(n) = \sum_nb(n) = \infty$ and $\sum_na(n)^2, \sum_nb(n)^2 < \infty$. 

First, let us define $F_{su}^\lambda(\Psi(j,b))$ and $M_{n+1}(s,u)$ such that:
\begin{align}
    F_{su}^\lambda (\Psi(j,b)) = (1-u) (R_0(s) + \lambda) + uR_1(s) + \gamma \sum_j p(j|i,u) \max_{v \in \{0,1\}} \Psi(j,v) \label{eq_F_su}
    \\
    M_{n+1}(s,u) = (1-u)(R_0(s) + \lambda_n(x)) + uR_1(s) + \max_{v \in \{0,1\}} Q_n(x_{n+1},v) - F_{su}^{\lambda_n(x)}(Q_n)  \label{eq_M_su}
\end{align}

We can now rewrite the Equation (\ref{eq_q-learning with x}) as:
\begin{equation}
    Q_{n+1}^x (s,u) = Q_n^x(s,u) + \alpha(n) \left[F_{su}^{\lambda_n(x)}(Q_n) - Q_n + M_{n+1}(s,u) \right]
    \label{eq_Q-learning modified}
\end{equation}

Comparing equations (\ref{eq_q-learning with x}) and (\ref{eq_Q-learning modified}) we make the correspondence $a(n) = \alpha(n)$, $h(x_n,y_n) = F_{su}^{\lambda_n(x)}(Q_n) - Q_n$, where $x_n = Q_n$ and $y_n= \lambda_n$ are the $Q$-value and Whittle index estimate respectively and $M_{n+1}(s,u)$ is the martingale difference sequence $M_{n+1}^{(1)}$. On the other hand, equations (\ref{eq:learingW}) and (\ref{eq_update y}) correspond to  $b(n)=\beta(n)$, $g(x_n, y_n) = Q_n^x(x,1) - Q_n^x(x,0)$ and a martingale difference sequence $M_{n+1}^{(2)} = 0$.

As in \cite{borkar2008stochastic}, we will consider three necessary conditions for equations (\ref{eq_update x}) and (\ref{eq_update y}) to be stable and converge to their optimal values $x_n \rightarrow x^*$ and $y_n \rightarrow y^*$.
\begin{enumerate}
    \item[A1] $h$ and $g$ must be Lipschitz continuous.
    \item[A2] $\{M_n^{(1)}\}$ and $\{M_n^{(2)}\}$ are martingale difference sequences.
    \item[A3] $\{a(n)\}$ and $\{b(n)\}$ satisfy:
    \begin{itemize}
        \item $a(n) >0, b(n) > 0$
        \item $\sum_n a(n) = \sum_n b(n) = \infty$, $\sum_n (a(n)^2 + b(n)^2 < \infty$
        \item $\frac{b(n)}{a(n)} \rightarrow \infty$
    \end{itemize}
\end{enumerate}

The full proof of $\mathbf{A1}$ can be easily verified on page 687 of \cite{abounadi2001learning}. In our notation, $M_{n+1}(s,u)$
corresponds to $M_{n+1}^{(1)}$ and $M_{n+1}^{(2)}=0$,  thus satisfying condition $\mathbf{A2}$. Finally, $\mathbf{A3}$ is also verified given that
$\beta(n) = o (\alpha(n))$.

Let us assume that the equations (\ref{eq_q-learning with x}) and (\ref{eq:learingW}) are bounded. We will prove this condition later. First we will rewrite the equation for the calculation of the indices (\ref{eq:learingW}) as:
\begin{equation}
    \lambda_{n+1}(x) = \lambda_n(x) + \alpha(n) \left(\frac{\beta(n)}{\alpha(n)}\right) (Q_n^x(x,1) - Q_n^x(x,0)) \label{eq_whittle_modified}
\end{equation}

Let $\tau(n) = \sum_{m=0}^n\alpha(m), m \geq 0$. We define $\bar Q(t), \bar \lambda(t)$ as the interpolation of the trajectories of $Q_n^x$ and $\lambda_n(x)$ on each interval $[\tau(n), \tau(n+1)], n\geq 0$ as:
\begin{align}
    \bar Q(t) = Q(n) + \left(\frac{t - \tau(n)}{\tau(n+1) - \tau(n)} \right) (Q(n+1) - Q(n))
    \\
    \bar \lambda(t) = \lambda(n) + \left(\frac{t-\tau(n)}{\tau(n+1) - \tau(n)} \right) (\lambda(n+1) + \lambda(n))
\end{align}
\[
    t \in \left[\tau(n), \tau(n+1)\right]
\]
which track the asymptotic behavior of the coupled o.d.e.s
\[
    \dot Q(t) = h(Q(t), \lambda(t)), \dot \lambda = 0
\]
where the latter is a consequence of $\frac{\beta(n)}{\alpha(n)}\rightarrow 0$ in (\ref{eq_whittle_modified}). From the reference frame of $Q(t)$, $\lambda(\cdot)$ is a constant of value $\lambda'$. Because of this, the first o.d.e. becomes $\dot Q= h(Q(t), \lambda')$, which is well posed and bounded, and has an asymptotically stable equilibrium at $Q_\lambda^*$ (Theorem 3.4, p. 689 in \cite{abounadi2001learning}). This implies that $Q_n^x - Q_{\lambda_n}^* \rightarrow 0$ as $n\rightarrow \infty$. 

On the other hand, for $\lambda(t)$, let us consider a second trajectory on another time scale, such that:
\begin{equation}
\begin{aligned}
    \Tilde{\lambda}(t) = \lambda(n) + \left( \frac{t - \tau'(n)}{\tau'(n+1) - \tau'(n)} \right) (g(n+1) - g(n))
    \\
    t \in \left[\tau'(n), \tau'(n+1) \right], \tau'(n) = \sum_{m=0}^n \beta(m), n \geq 0
\end{aligned}
\end{equation}
which tracks the o.d.e.:
\[
    \dot \Lambda(t) = Q_{\Lambda(t)}^*(x,1) - Q_{\Lambda(t)}^*(x,0)
\]

If $\Lambda(t) > \lambda(x)$ (excess subsidy), the passive mode is preferred, i.e., $Q_{\Lambda(t)}^*(x,0) > Q_{\Lambda(t)}^*(x,1)$, making the r.h.s of the previous equation $<0$ and $\Lambda(t)$ decreases. Likewise, if the opposite strict inequality holds, the r.h.s. is $>0$ and $\lambda(t)$ increases. Thus, the trajectory of $\Lambda(\cdot)$ remains bounded. As in the previous case, being a well-defined  bounded scalar o.d.e., it converges to an asymptotically stable equilibrium  where $\Lambda$ satisfies $Q_\Lambda^*(x,1) = Q_\Lambda^*(x,0)$. This point is where both policies are equally desirable, i.e. $\Lambda$ is the Whittle index.

\end{document}